\documentclass[twoside,11pt]{article}
\usepackage{jair, rawfonts}

\usepackage{url}

\usepackage{graphicx} 
\usepackage{amsmath}
\usepackage{amsthm}
\usepackage{subcaption}
\usepackage{amssymb}
\usepackage{xcolor}
\usepackage{algorithm}
\usepackage[noend]{algcompatible}
\usepackage{cleveref}
\usepackage[style=apa,natbib=true,doi=false,url=false]{biblatex}
\ShortHeadings{STT-CBS: A Conflict-Based Search Algorithm}
{Peltzer, Brown, Schwager, Kochenderfer, \& Sehr}
\firstpageno{1}

\begin{document}

\title{STT-CBS: A Conflict-Based Search Algorithm for Multi-Agent Path Finding with Stochastic Travel Times}

\author{\name Oriana Peltzer \email peltzer@stanford.edu \\
       \name Kyle Brown \email kylejbrown17@gmail.com \\
       \addr Mechanical Engineering, Stanford University, CA 94305 USA
       \AND
       \name Mac Schwager \email schwager@stanford.edu \\
       \name Mykel J. Kochenderfer \email mykel@stanford.edu \\
       \addr Aeronautics and Astronautics, 
       Stanford University, CA 94305 USA
       \AND
       \name Martin A. Sehr \email martin.sehr@getcruise.com \\
       \addr Cruise Automation, San Francisco, CA 94107 USA}

% For research notes, remove the comment character in the line below.
% \researchnote

\maketitle

\begin{abstract}
We present an algorithm for finding optimal paths for multiple stochastic agents in a graph to reach their destinations with a user-specified maximum pairwise collision probability.  Our algorithm, called STT-CBS, uses Conflict-Based Search (CBS) with a stochastic travel time (STT) model for the agents.  We model robot travel time along each edge of the graph by independent gamma-distributed random variables, and propose probabilistic collision identification and constraint creation methods to robustly handle travel time uncertainty.  We show that under reasonable assumptions our algorithm is optimal in terms of expected sum of travel times, while ensuring an upper bound on each pairwise conflict probability. Simulations and hardware experiments show that STT-CBS is able to significantly decrease conflict probability over CBS, while remaining within the same complexity class.
\end{abstract}

% --------------------- INTRO --------------------- %
\section{Introduction}\label{sec:introduction}

% ----------------- INTRODUCTION ---------------- %

% Motivation
We are interested in routing a team of robots from an initial configuration to a target configuration through collision-free trajectories, represented as paths through a graph.  Relevant applications include real-time vehicle routing \citep{routingstochastictt}, warehouse management \citep{warehousemanagement}, and unmanned aerial vehicle (UAV) coordination \citep{uav}.  Typically in such applications, routes are planned offline for the whole team assuming known, deterministic nominal travel times for robots to traverse each node and edge in the graph. The planned paths are then executed by the robots through lower-level path following controllers.  Unfortunately, real robots are stochastic.  Hence the actual travel times for robots quickly depart from the nominal expected times in the plan, leading to potential unplanned robot-robot conflicts.  If robots collide with one another, or need to take unplanned-for maneuvers to avoid collisions, this leads to further delays.  This causes a cascading effect, making their travel times vary more widely from what was expected during planning, and leading to more potential downstream robot-robot conflicts.  The consequent lack of predictability makes optimal planning on long time horizons challenging. To address this problem, we propose Stochastic Travel Time Conflict Based Search (STT-CBS), a planning algorithm that inherently plans for stochasticity in robots' node and edge traversal times.  STT-CBS minimizes robots' expected travel time subject to a pairwise conflict probability among the robots.% compared to the state-of-the-art.

% MAPF
We build upon the common Multi-Agent Path Finding (MAPF) problem formulation, wherein agents are allowed to move along the edges of a graph from their initial locations to prescribed destinations.
%Current approaches use deterministic travel time
Recently, optimal MAPF planners have shown their effectiveness empirically on a variety of problems \citep{cbs,pushnrotate,cbslargeagents}, most of which assume the robots travel along edges in the graph with fixed, deterministic travel times.
% Main advantage of considering uncertainty
In contrast, in this work we focus on accounting for stochasticity in the robots' travel times.  This is important, as real robots (e.g., in a factory, or a warehouse) may not traverse an edge at exactly the intended rate.  Real robots have to contend with wheel-slip, course-correction and collision avoidance actions, battery voltage fluctuations, and many other effects that randomly influence the time it takes them to traverse nodes and edges in the graph.  If one ignores these random effects, the resulting trajectories may lead to collisions, and may yield significantly suboptimal total travel times in practice.  

% Contribution
Our proposed algorithm, STT-CBS, is a MAPF planning algorithm that seeks to minimize the expected sum of travel times of the robots subject to a bound on the probability of collision for any pair of robots. By accounting for uncertainty in travel time due to environmental disturbances, STT-CBS improves the consistency and reliability in the actual execution of the robots' planned paths.
%, leading to \textcolor{blue}{XX\%} lower expected total travel time, and \textcolor{blue}{XX\%} lower probability of collision compared with standard CBS \citep{cbs}.
%Model
We model the time that each robot waits at a node of the graph 
as a gamma-distributed random variable. This model of randomness captures the sequential summing of uncertain effects along a trajectory as a robot passes through multiple nodes in the graph.  We only ascribe random delays to the nodes in our model.  Conceptually, we lump the delay due to edge traversal and node traversal into one random effect at the nodes only.  We suppose it is more important to model random delays at the nodes, since they act as ``intersections'' in the road map, and hence are likely to be key choke points where robot-robot conflicts might occur.  One can also ascribe separate random delays to the traversal of edges, but with an increased computational burden in the algorithm.

% We model the travel time for each robot to traverse an edge of the graph\footnote{One can equivalently associate the delay with the edge or either of the two nodes for that edge.  Without loss of generality, we choose to associate the delay with the end node rather than the edge itself.} % 
%as a gamma-distributed random variable. This model of randomness captures the sequential summing of uncertain effects along a trajectory of multiple edge traversals. 

%Properties
We show that if a solution exists, our algorithm will return the solution that minimizes expected total travel time subject to the probabilistic collision constraint. 
%Simulations and experiments
We compare STT-CBS against a %state-of-the-art
baseline algorithm on a set of simulation experiments, showing that our algorithm can better prevent collisions under this stochastic travel time model.  We also demonstrate our algorithm in hardware experiments, routing three ground robots from initial locations, through a graph, to goal locations in a lab environment, while accounting for the real-world stochasticity in their travel times.
%Original
%To describe the uncertainty due to environmental disturbances, we model the travel time for each robot to traverse an edge of the graph\footnote{One can equivalently associate the delay with the edge $e_{ij}$ or either of the two nodes for that edge, $v_i$ or $v_j$.  Without loss of generality, we choose to associate the delay with the end node $v_j$ rather than the edge itself.} as a gamma-distributed random variable.  This model of randomness captures the sequential summing of uncertain effects along a trajectory of multiple edge traversals.  Under this model, we propose Stochastic Travel Time Conflict-Based Search (STT-CBS), a novel MAPF planning algorithm that seeks to minimize the sum of travel times of the robots subject to a bound on the probability of collision for any pair of robots.  The algorithm reasons explicitly about stochasticity in travel time as robots move along a path. 
%We show that if a solution exists, our algorithm will return the solution that minimizes expected total travel time subject to the probabilistic collision constraint. We compare STT-CBS against the original CBS algorithm on a set of experiments, showing that our algorithm can better prevent collisions under this stochastic travel time model.
% ----------------------------------------

\section{Background and Related Work}

% Discrete-time: pebble motion
%\textcolor{blue}{Mac: I would weave details about our algorithm throughout this background section.  The reader wants to know how is STT-CBS different from existing methods? what does it take from existing methods? what assumptions are being made in STT-CBS and how are they different from standar assumptions?} 
In the discrete-time MAPF formulation, edges can be traversed in unit time, and each robot is assigned to a specific goal.  The unit travel time is sometimes called the ``pebble motion'' problem \citep{Korn}, and each agent having an assigned goal is called the ``labeled'' case.  The solution to a discrete-time MAPF problem is a joint assignment of trajectories such that no two agents occupy the same vertex or traverse the same edge at the same time (i.e., trajectories are free of conflict).  
This model is the basis of many graph-based multi-agent path finding algorithms \citep{cbs,umstar,flow}.  However, robots often do not satisfy the strict assumptions of the pebble motion model. Recent work in MAPF has focued on more realistic robot models, such as continuous-time edge traversal \citep{smtcbs} and probabilistic integer delays at nodes \citep{probustcbs,umstar}.  Our algorithm, STT-CBS, is based on a more general model, where travel time for the robots is modeled as a continuous gamma-distributed random variable.
One may choose to optimize a variety of different objective functions for MAPF problems, however the most common choices are to minimize total travel time (the sum of the lengths of all trajectories) \citep{cbs,flow}, or to minimize makespan (the travel time of the longest robot trajectory) \citep{kyle}.  In keeping with most of the literature, we choose to minimize total travel time, although our algorithm could be adapted to minimize makespan or other objectives.  

Finding an optimal solution to the labeled discrete-time MAPF problem is NP-hard \citep{nphard}.
% List of optimal and complete solution methods
Optimal and \emph{complete} MAPF solution methods (those that are guaranteed to find an optimal solution if it exists) can be broadly grouped into three categories: 
In \emph{(i) coupled approaches}, the search for an optimal feasible solution is performed directly in the joint action space of the agents. 
\emph{(ii) Decoupled approaches} operate by computing paths on an individual (per-agent) basis, and then resolving pairwise conflicts repeatedly by re-planning. This is the basis of Conflict-Based Search \citep{cbs} and its variants \citep{cbs}, as well as the algorithm we propose in this paper, STT-CBS.
There are also \emph{(iii) semi-coupled approaches}, which group agents into teams or ``meta-agents'', and then apply a de-coupled approach %(e.g. conflict-based search) 
between the joint plans associated with each team. A representative example of a semi-coupled approach is Operator Decomposition and Independence Detection (OD+ID) \citep{odid}.

\begin{figure}[ht]
\centering

  \includegraphics[scale=0.25]{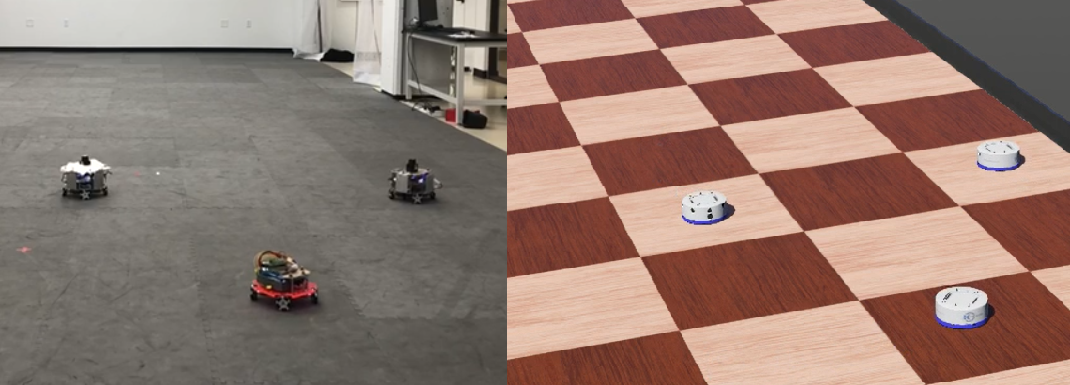}
  \caption{Our algorithm plans collision-free paths in a graph for robots that travel with stochastic travel times through a path in a graph. We evaluated the algorithm in hardware and simulation experiments with robots to verify the robustness of our approach to real-world stochastic delays.}
  \label{fig:intro_bots}

\end{figure}

The ``pebble-motion'' MAPF formulation abstracts away robot dynamics and does not account for uncertainty.  In trying to execute paths specified by a MAPF planner, actuator dynamics and other disturbances will inevitably introduce errors that violate the assumptions of constant-velocity and unit-time edge traversal. This can be problematic when nominally ``conflict-free'' trajectories are not conflict-free during actual execution. 
One way to address this modeling error is to ignore it during planning, and then to incorporate closed-loop control for stabilizing robots to the commanded trajectories \citep{cocktailapproach}. While this approach is simple and practical, it relies on important spacing around agents in order to avoid accumulating delays and deadlocks. It provides no means to predict or assess conflict probabilities and associated delays that may occur due to low-level controllers taking collision avoidance maneuvers. In automated warehouse and manufacturing applications, space efficiency, time efficiency, and predictability are essential \citep{warehousedesignoptimization}. To this end, we propose a solution that is able to perform more consistently in a cluttered environment, while allowing for a user to specify maximum pairwise conflict probability between robots.
%The effectiveness of such an approach depends heavily on the types of disturbances that affect the execution.

Our algorithm builds upon Conflict-Based Search (CBS) \citep{cbs}, a state-of-the-art algorithm to solve the optimal MAPF problem. Many heuristics \citep{cbsheuristics,disjointcbs} have been developed to accelerate CBS. It has also been merged with different planning methods \citep{cbm} and extended to more complex agent geometries \citep{cbslargeagents} and graphs that allow for non-unit edge traversal times \citep{smtcbs}.
% All suffer from the same problem as CBS, namely that the final solution is not necessarily robust to conflicts that might occur due to unmodeled delay in the actual execution of those paths.
None of these approaches are robust to conflicts that might occur due to delay in the actual execution of the robots' paths. Stochastic travel times are important, since what may start as a small delay by one robot may snowball into a traffic jam that significantly delays all robots.  
Few MAPF approaches reason about robots' execution uncertainty during the planning phase. 
\citet{shahar} propose an algorithm to compute optimal conflict-free paths in worst-case scenarios under bounded travel times. While this model is more general than the traditional pebble-motion model, it fails to account for the probability of conflicts occurring. Thus, solutions where a highly unlikely travel time combination within the specified bounds results in a conflict are discarded, which makes the choice of conservative upper bounds particularly costly. 
Another algorithm that does reason about uncertain travel times is $k$-robust CBS, which seeks solutions that remain collision free considering that each agent may be delayed for up to $k$ time steps \citep{robustcbs}. A downside of this model is that it fails to consider the sequential summing of delay as agents advance, which is key to capturing the snowballing effect of a traffic jam among the robots.
In order to take this summing into account, Atzmon et al. have extended $k$-robust CBS to $p$-robust CBS \citep{probustcbs}. The pR-CBS algorithm searches for solutions minimizing total travel time such that the probability of at least one conflict occurring during path execution is smaller than a threshold $p$. In this model, agents have a fixed probability of being delayed at their current node for one time step. This approach iterates over all possible solutions sorted by increasing cost until one is found that is feasible and has an overall conflict probability that is smaller than $p$.
Alternatively, the UM$^\star$ algorithm \citep{umstar} also takes sequential summing of delay into account. In this approach, progress is modeled as a Markov Decision Process. This algorithm fits in the framework of M$^\star$ \citep{mstar}, and provides an optimal planning solution for the MAPF problem within the pebble motion model. 
While the UM$^\star$ algorithm is incomplete, it is empirically effective to compute optimal solutions for a large category of problems.
Approximate Minimization in Expectation \citep{delayprobabilities}, based on the same model, computes solutions that approximately minimize expected makespan. One advantage of this method compared to the previous is that it takes into account the possibility that agents may collide with each other during path execution.

Our proposed algorithm, STT-CBS, similarly plans under stochastic travel times. However, it does not require the unrealistic assumptions within the pebble-motion model. Specifically, our model allows for the robots to have different travel speeds, accommodates non-unit time traversal of edges and nodes, and allows for gamma-distributed stochastic delays at nodes, where different nodes can have different parameters for their gamma distributions. Allowing for different robot speeds can better reflect the nominal time that robots take to navigate when accounting for surrounding obstacles, and the gamma-distributed delay at different nodes may be adjusted to differentiate areas where sources of uncertainty, such as obstacles or people, are more or less present. 

 %However, it does not require the assumptions required for the pebble-motion model, where each edge should be traversed in unit time, and nodes should be placed close enough spatially such that only one robot can occupy a node at one time step. This model is useful when robots are identical and navigate at the same constant velocity, thus ignoring the effect of turns. 
 %By accounting for uncertainty in travel time due to environmental disturbances, STT-CBS improves the consistency and reliability in the actual execution of the robots' planned paths, leading to lower expected total travel time, and lower probability of collision.

% ------------------------------------------------- %

% --------------------- PROBLEM ------------------------ %
\section{Approach}\label{sec:prbblem_statement}

Here we formalize the stochastic travel time MAPF problem under consideration. As in the deterministic travel time formulation, each agent must move from its initial location to a prescribed final location without collision with other agents. The task is to find a solution that minimizes the \emph{expected} total travel time subject to the constraint that the likelihood of collision for each pair of agents at any place in the graph is below some threshold $\epsilon \in [0,1]$.  While we present the algorithm as an offline planner for clarity, in practice one could re-solve for the agents' trajectories periodically during execution in a receding horizon fashion, in order to incorporate real time knowledge of the agents' locations and delays up to that point.

%The solution generated by STT-CBS is an open-loop motion plan.
    %While this kind of planner would likely be combined with closed-loop feedback control in practice, we are most interested in comparing the open-loop plan with that produced by Conflict-Based Search.
%}
% \RobotPos{<idx>}{<time>}
\newcommand{\RobotPos}[2]{p_{#1}(#2)}
% \TimeDuration{<idx>}{<time>}
% \newcommand{\TimeDuration}[2]{\tau_{}}
% \newcommand{\Agent}[1]{A_{#1}}
\newcommand{\Agent}[1]{\text{Robot } {#1}}
\newcommand{\Path}[1]{P_{#1}}
\newcommand{\Node}{v}
\newcommand{\NodeN}[1]{v_{#1}}

% \TotalTraversalTime{<agent>}{<idx>}
\newcommand{\TotalTraversalTime}[2]{\tau_{#1}(#2)}

% \NominalArrivalTime{<agent>}
\newcommand{\NominalArrivalTime}[1]{t_{#1}}
\newcommand{\NominalEdgeTravelTime}{t_e}

% \CumulativeDelay{<agent>}{<idx>}
\newcommand{\CumulativeDelayLong}[2]{\Delta_{#1}(#2)}
\newcommand{\CumulativeDelay}[1]{\Delta_{#1}}
% \DelayAtNode{<node>}{<agent>}
\newcommand{\DelayAtNode}[2]{\tau_{#1, #2}}

\subsection{Problem Setup} We consider a team of robots simultaneously traversing paths on a graph $\mathcal{G} = (\mathcal{V}, \mathcal{E})$, where $\mathcal{V} = \{1, \ldots, N\}$ is a set of nodes of the graph and $\mathcal{E} = \{\ldots, (i,j), \ldots\}$ is a set of edges. We denote the discrete path taken by robot $i$ through the graph by the tuple $\Path{i} = (\RobotPos{i}{0}, \RobotPos{i}{1}, \ldots, \RobotPos{i}{K_i})$, with discrete path length $K_i$, and we require that all neighboring nodes in this tuple are linked by edges in the graph, $e_i(k) = (\RobotPos{i}{k-1},\RobotPos{i}{k})\in\mathcal{E}$.  Physically, the nodes in the graph represent points in the environment, and the edges represent physical paths between those points, which require some finite time for the robot to traverse. Each edge in the graph has a travel time $\NominalEdgeTravelTime$, and if robot $i$ traverses edge $e$ at time $k$, we say $t_{e_i}(k) = \NominalEdgeTravelTime$.  We model the time that a robot $i$ waits at the $k$th node in its path as a real-valued time duration $\tau_i(k)$.  The robot may need to wait at a node to avoid a collision, while another robot passes through neighboring edges or vertices.  We model this waiting time with a gamma distributed random variable.  One can also ascribe random travel times to the robots' edge traversals, although we do not do so in this work.  Thus, the history of time durations for the path of robot $i$ is given by both the sequence of deterministic edge traversal times $T_{ei} = (t_{e_i}(1), \ldots, t_{e_i}(K_i))$ and stochastic vertex traversal times $T_{vi} = (\tau_i(0), \ldots, \tau_i(K_i))$.  Together, $\Path{i}$, $T_{ei}$ and $T_{vi}$ define the \emph{trajectory} of robot $i$.

\subsection{Uncertainty Model}\label{sec:uncertainty_model}

%The choice of uncertainty model is highly dependent on the target application.
    %In UAV applications, for example, we might be concerned about the effect of wind blowing the robots off of the desired path.
    In warehouse or factory-floor settings, robots have low-level controllers that control them to track a desired path.  However, disturbances are likely to occur as delays along the prescribed path due to wheel slip, feedback corrections from the low-level controller, unexpected minor obstacles, and other sources of stochastic delay.  Our particular choice of travel time model is motivated by this phenomenon of stochastic accumulating delay for factory or warehouse robots.
    
We model delay at each node of an agent's trajectory as a random variable following a gamma distribution. More precisely, we consider that each time a node $\Node$ %(or edge $e$)% 
is traversed by $\Agent{i}$, then the robot is delayed, and the delay $\DelayAtNode{v}{i}$ is a random variable described by $\DelayAtNode{v}{i} \sim \text{Gamma}(n_v,\lambda)$. The more nodes and edges traversed by each agent, the larger its expected delay and uncertainty in its position along its route in the graph.  %As described above, the choice to associate the delay to a node rather than an edge is arbitrary.
% With the model described, we are able to encompass interactions in between robots at intersections, where they are the most likely to be delayed. 

\subsection{Computing Conflict Probability}\label{sec:conflict_probability}

\begin{figure*}
\centering
\includegraphics[scale=0.5]{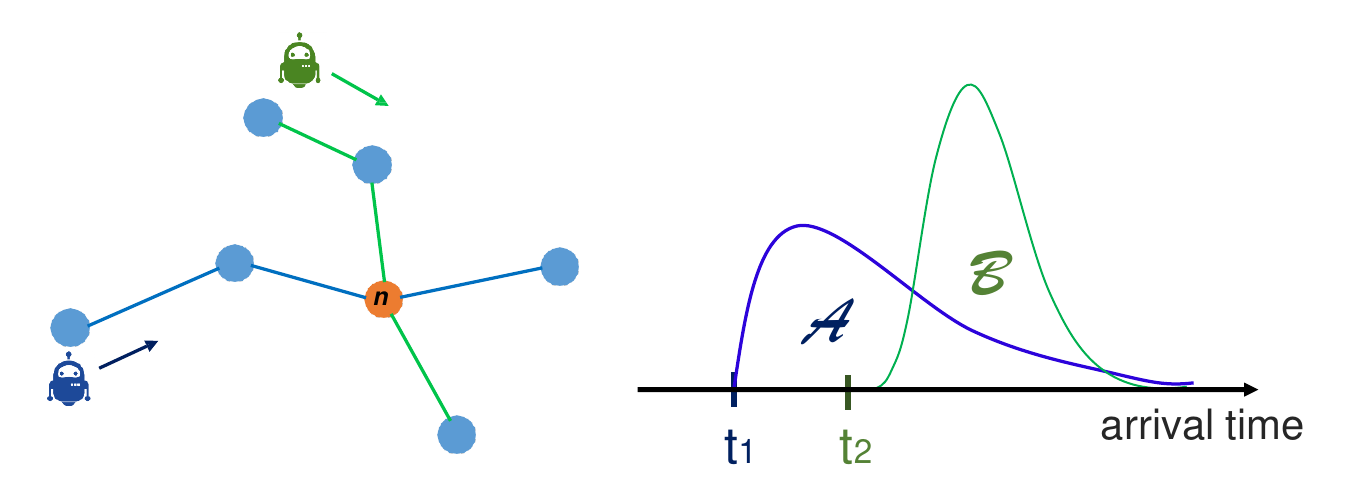}
\caption{Conflict identification at a node $n$. Blue and green robots are scheduled to arrive at node $n$ at different times, $t_1$ and $t_2$. Distributions $\mathcal{A}$ and $\mathcal{B}$ illustrate the probability density of the blue and green robot's arrival time respectively. %Their expected arrival times, accounting for stochastic delays, are $t_2$ and $t_4$ respectively. 
%\textcolor{blue}{Mac: Sorry, but this caption isn't clear.  $t_1$ is the arrival time and $t_2$ the departure time for the blue robot?  What is a "traversal time" versus an "arrival time"?  Where is the departure time?  Also, shouldn't the distribution for the green robot be drawn in green, not red?  Please revise.} 
Although their arrival times are different, uncertainty over travel time may cause them occupy node $n$ at overlapping time intervals, causing a conflict. If the green robot arrives at $n$ before the blue robot departs from $n$, and if the blue robot arrives before the green robot departs, then the robots will conflict with each other.}
\label{fig:delay}
\end{figure*}

In this section, we express the probability of a conflict between two agents as a function of our model parameters.
\Cref{fig:delay} illustrates how a conflict can happen in this stochastic model.

%\vspace{2mm}

\subsubsection{Cumulative delay}

% Let \Agent{i} be an agent with an assigned path \Path{i}. At the start of its route, it is not delayed. As it progresses along its route, it is delayed at each node $\RobotPos{i}{k}$ (or edge $e_i(k)$) by some value $\nu_{\RobotPos{i}{k}}$, with $\nu_{\RobotPos{i}{k}} \sim \text{Gamma}(n_{\RobotPos{i}{k}},\lambda)$ (the same holds for edge $e_i(k)$). We note the total delay accumulated by agent $Ai$ at its $k$th node as $\nu_i(k)$.

Let $\Agent{i}$ be assigned a path $\Path{i}$. At the start of its route, it is not delayed. As it progresses along its route, it is delayed at each node $\Node$ by some value $\tau_i(k)$ in addition to its planned travel time, with $\tau_i(k) \sim \text{Gamma}(n_{v},\lambda)$. We note the total delay accumulated by $\Agent{i}$ when it arrives to the $k$th node on its path as $\CumulativeDelayLong{i}{k} =\sum_0^k\tau_i(k)$.

% In summary, we can express the time duration $\TotalTraversalTime{i}{k}$ as the sum of nominal edge traversal time and the delay experienced by the agent at the first node. Therefore, we can express the time duration
% \begin{equation}
% \label{traveltime}
%     \TotalTraversalTime{i}{k} = t_{e_i(k)} + \DelayAtNode{\RobotPos{i}{k}}{i}
% \end{equation}
% and the total travel time
% \begin{equation}
% \label{totaltraveltime}
% \begin{aligned}
%     \sum_{j=1}^{k} \TotalTraversalTime{i}{j} &= \sum_{j=1}^{k} (t_{e_i(j)} + \DelayAtNode{\RobotPos{i}{j-1}}{i}) \\
%     &= t_i(k) + \CumulativeDelayLong{i}{k}
%     ,
% \end{aligned}
% \end{equation}
% where $e_i(j)$ is the edge $(\RobotPos{i}{j}, \RobotPos{i}{j+1})$ along the robot's path, and $t_i(k)$ is the nominal arrival time of the agent at position $\RobotPos{i}{k}$.

For simplicity, we use a rate parameter $\lambda$ in the Gamma distribution that is identical across all the nodes in our graph. %, although our method can accommodate heterogeneous $\lambda$ values, as well. 
To account for different travel time distributions, we can vary the shape parameter $n_v$.
%\textcolor{blue}{Mac: Please clarify whether this is a convenience or it is required.  If just a convenience, please explain how one can deal with different $\lambda$.}.
Conveniently, the sum of $N$ independent gamma-distributed random variables with shape parameters $n_v$ and identical rate parameter $\lambda$ is a gamma-distributed random variable with shape parameter $\sum_{v=1}^{N}n_v$ and rate parameter $\lambda$.
Thus we can write the delay accumulated by $\Agent{i}$ along its trajectory up to time $k$ as

\begin{equation}
\label{cumulative_delay}
    \CumulativeDelayLong{i}{k} \sim \text{Gamma}\left(\sum_{j=0}^{k}n_{\RobotPos{i}{j}},\lambda\right).
\end{equation}

Using this stochastic delay, we can express the probability of a conflict between two robots, $\Agent{1}$ and $\Agent{2}$, at a particular node or edge. Dropping the index $k$ and defining $n_1 = \sum_{j = 0}^kn_{p_1(j)}$ and $n_2 = \sum_{j = 0}^kn_{p_2(j)}$ for notational simplicity, we will denote the cumulative delays for each agent
$\CumulativeDelay{1} \sim \text{Gamma}(n_1,\lambda)$ and $\CumulativeDelay{2} \sim \text{Gamma}(n_2,\lambda)$. %, and can similarly express the nominal travel times as $\NominalArrivalTime{1}$ and $\NominalArrivalTime{2}$.

\subsubsection{Node Conflicts}\label{sec:node_conflicts}
The probability of a conflict between $\Agent{1}$ and $\Agent{2}$ at node $\Node$ is computed as follows.
Let $\NominalArrivalTime{1}$ and $\NominalArrivalTime{2}$ denote the nominal times (as computed without any stochastic delays) at which $\Agent{1}$ and $\Agent{2}$ are scheduled to arrive at $\Node$, respectively.
The actual arrival times depend on the agents' respective accumulated delays, $\CumulativeDelay{1}$ and $\CumulativeDelay{2}$, which are distributed according to \cref{cumulative_delay}. The agents may also accumulate additional delays---denoted by the independent random variables $\DelayAtNode{v}{1}$ and $\DelayAtNode{v}{2}$, respectively---before leaving $\Node$.

% t_{A1} -> \NominalArrivalTime{1}
% \nu_{R1} -> \CumulativeDelay{1}
% \nu_{D1} -> \DelayAtNode{v}{1}

$\Agent{1}$ and $\Agent{2}$ will conflict with each other if there exists a time interval during which they are both present at node $\Node$. This occurs if and only if the following two events occur: ($A$) ``$\Agent{1}$ leaves after $\Agent{2}$ arrives'', and ($B$) ``$\Agent{2}$ leaves after $\Agent{1}$ arrives.''
% \begin{itemize}
%     \item $A$: ``agent R1 leaves after agent R2 arrives''
%     \item $B$: ``agent R2 leaves after agent R1 arrives.''
% \end{itemize}

\vspace{1mm}
A conflict can therefore be formally defined as

\begin{equation*}
  \begin{aligned}
    \text{Conflict} & \Leftrightarrow
        \begin{cases}
        \NominalArrivalTime{1} + \CumulativeDelay{1} + \DelayAtNode{v}{1} \geq \NominalArrivalTime{2} + \CumulativeDelay{2} \quad (A)
        \\
        \NominalArrivalTime{2} + \CumulativeDelay{2} + \DelayAtNode{v}{2} \geq \NominalArrivalTime{1} + \CumulativeDelay{1} \quad(B)
        \end{cases} \\
     & \Leftrightarrow
        \begin{cases}
        \CumulativeDelay{1} - \CumulativeDelay{2} \geq  \NominalArrivalTime{2} - \NominalArrivalTime{1} - \DelayAtNode{v}{1} \quad (A)
        \\
        \CumulativeDelay{1} - \CumulativeDelay{2} \leq \NominalArrivalTime{2} - \NominalArrivalTime{1} + \DelayAtNode{v}{2} \quad (B)
        \end{cases}.
  \end{aligned}
\end{equation*}

Let $y = \CumulativeDelay{1} - \CumulativeDelay{2}$ be the difference in accumulated delays between both robots. Because $\DelayAtNode{v}{1}$ and $\DelayAtNode{v}{2}$ are independent, $A$ and $B$ are conditionally independent given $y$. Thus we can write
%\textcolor{blue}{Mac: What is $y$, and why are these conditionally independent?  Please give both some intuition and mathematical reasoning.}
$$
P(\text{Conflict}) = P(A\cap B) = \int_{y} P(A \mid y)P(B \mid y)P(y) dy
.$$
Writing $\text{Gamma}(x \mid n,\lambda)$ as $G(x \mid n,\lambda)$, we have
\begin{equation*}
    \begin{aligned}
    P(A \mid y) &= P(\DelayAtNode{v}{1} \geq  \NominalArrivalTime{2} - \NominalArrivalTime{1} - y) 
    &= \int_{(\NominalArrivalTime{2} - \NominalArrivalTime{1} - y)^+}^{\infty} G(x \mid n_{\Node}, \lambda)dx ,
    \end{aligned}
\end{equation*}
and
\begin{equation*}
    \begin{aligned}
    P(B \mid y) &= P(\DelayAtNode{v}{2} \geq  \NominalArrivalTime{1} - \NominalArrivalTime{2} + y)
    &= \int_{(\NominalArrivalTime{1} - \NominalArrivalTime{2} + y)^+}^{\infty} G(x \mid n_{\Node}, \lambda)dx,
    \end{aligned}
\end{equation*}

where $(\cdot)^+ = \max\{0, \cdot\}$.  We notice that for fixed $\NominalArrivalTime{1}$, $\NominalArrivalTime{2}$, and $y$, at least one of the two quantities $(\NominalArrivalTime{2} - \NominalArrivalTime{1} - y)^+$ or $(\NominalArrivalTime{1} - \NominalArrivalTime{2} + y)^+$ will be equal to 0. Therefore, at least one of $P(A \mid y)$ or $P(B \mid y)$ will be equal to 1. %Intuitively, this is due to the fact that no matter the outcome, at least one of the events $\mathcal{A}$ or $\mathcal{B}$ is always true. 
Simplifying the product
$$
P(A \mid y) P(B \mid y) = \int_{\vert \NominalArrivalTime{1} - \NominalArrivalTime{2} + y \vert}^{\infty} G(x \mid n_{\Node}, \lambda )dx ,
$$
and expressing $P(y)$ as 
\begin{equation*}
\begin{aligned}
    P_{neg}(y) &= \int_{0}^\infty G(t \mid n_1, \lambda)G(t-y \mid n_2, \lambda) dt, \\
    P_{pos}(y) &= \int_{0}^\infty G(y+t \mid n_1, \lambda)G(t \mid n_2, \lambda) dt, \\
    P(y) &= \begin{cases}
        P_{neg}(y) \quad \mbox{if} \quad y \leq 0 \\
        P_{pos}(y) \quad \mbox{if} \quad y > 0 ,
    \end{cases}
\end{aligned}
\end{equation*}
we can finally write that for a vertex $\Node$,
\begin{equation}
\label{nodeprobability}
    \begin{aligned}
        P(\text{Conflict}) = \int_{0}^{\infty} P_{pos}(y)\int_{\vert \NominalArrivalTime{2} - \NominalArrivalTime{1} - y \vert}^\infty G(x \mid n_{\Node}, \lambda)dx dy 
        \\
        + \int_{-\infty}^{0} P_{neg}(y')\int_{\vert \NominalArrivalTime{2} - \NominalArrivalTime{1} - y' \vert}^\infty G(x'\mid n_{\Node}, \lambda)dx' dy' .
    \end{aligned}
\end{equation}
We will denote this function $P(\text{Conflict}) = P_{c\Node}(\NominalArrivalTime{1}-\NominalArrivalTime{2},n_1,n_2,\lambda,n_{\Node})$.

There are multiple ways to compute the probability from \cref{nodeprobability}, the simplest being Monte Carlo simulation. It is also possible to numerically integrate the density over variable $y$, e.g., by quadrature. We chose in our experiments to use Monte Carlo simulation for the low variability of its computation time compared to other numerical integration methods. %The latter option also requires a careful choice of parameters and integration bounds.
%\textcolor{blue}{Mac: can we actually write this whole section as a Theorem/proof?  I think that would make it stronger and better organized.  We can also claim this as a contribution in the abstract/intro.} 
%Using \cref{nodeprobability}, we can numerically integrate the density over variable $y$. In order to speed up computation, we integrate in a smaller region surrounding the mode (which can be computed analytically) of this distribution. \textcolor{blue}{Mac: This requires more explanation.  What did we actually do, why, and how well does it work compared to other integration methods? What exactly is the meaning of a "smaller region," here, and how can it be computed analytically (what is the formula, or where can it be found?).  Is it approximate?  Do we compute to ensure conservatism?  These are all questions we need to answer precisely here.  We are too vague here.} 
%It is also possible to evaluate \cref{nodeprobability} using Monte Carlo simulation. 
% of this distribution, which we are able to derive analytically. 

\vspace{2mm}

\subsubsection{Edge Conflicts}\label{sec:edge_conflicts}

While we do not model the edge traversal as adding a random delay, the arrival and departure times of a robot traversing an edge are still random, due to the random delay accumulated at the nodes preceding the edge traversal.  Therefore, we also must consider the probability of conflicts at edges.  %\textcolor{blue}{Mac: Please update the notation in this section to stylistically match the other sections.  E.g., $v_1$ and $v_2$ instead of $N1$ and $N2$, Robot 1 instead of $R1$, etc.}  
Let $e$ denote the edge between nodes $\NodeN{1}$ and $\NodeN{2}$. Let us suppose that $\Agent{1}$ traverses the edge from $\NodeN{1}$ to $\NodeN{2}$, $\Agent{2}$ traverses from $\NodeN{2}$ to $\NodeN{1}$, and that the edge takes time $t_e$ to traverse.
Let $\NominalArrivalTime{1}$ and $\NominalArrivalTime{2}$ denote the nominal scheduled departure times of robot $\Agent{1}$ from node $\NodeN{1}$ and robot $\Agent{2}$ from node $\NodeN{2}$, respectively.
With $\CumulativeDelay{1}$ as the delay accumulated by $\Agent{1}$ until node $\NodeN{1}$ and $\CumulativeDelay{2}$ as the delay accumulated by $\Agent{2}$ until $\NodeN{2}$, we can similarly express events $A$ and $B$ for an edge conflict as
\begin{equation*}
  \begin{aligned}
    \text{Conflict} & \Leftrightarrow
        \begin{cases}
        \NominalArrivalTime{1} + \CumulativeDelay{1} + t_e \geq \NominalArrivalTime{2} + \CumulativeDelay{2} \quad (A)
        \\
        \NominalArrivalTime{2} + \CumulativeDelay{2} + t_e \geq \NominalArrivalTime{1} + \CumulativeDelay{1} \quad (B)
        \end{cases} \\
     & \Leftrightarrow
        \begin{cases}
        \CumulativeDelay{1} \geq \CumulativeDelay{2} + \NominalArrivalTime{2} - \NominalArrivalTime{1} - t_e \quad (A)
        \\
        \CumulativeDelay{1} \leq \CumulativeDelay{2} + \NominalArrivalTime{2} - \NominalArrivalTime{1} + t_e \quad (B)
        \end{cases}.
  \end{aligned}
\end{equation*}
Defining $b = x +\NominalArrivalTime{2}-\NominalArrivalTime{1}$, and simplifying, we have that on an edge $e$,
\begin{equation}
\label{edgeprobability}
    \begin{aligned}
    P(\text{Conflict})
    &= P_{ce}(\NominalArrivalTime{1}-\NominalArrivalTime{2},n_1,n_2,\lambda,t_e) \\
    &= \int_{0}^{\infty} G(x \mid n_1, \lambda) \int_{(b-t_e)^+}^{(b+t_e)^+} G(w \mid n_2, \lambda) dw dx .
    \end{aligned}
\end{equation}

Using the cumulative density function for the gamma distribution, this expression only needs to be integrated over variable $e_2$.

% ------ old remark (removed) ------
%\subsubsection{Remark 1.}

%\vspace{2mm}
%%Is this useful?
%Notice that under our model, computing the probability of conflicts on edges is relatively inexpensive compared to conflicts on nodes. Using the cumulative density function for the gamma distribution, the former only implies to integrate over one variable, here $e_2$. The reason for this is that we do not introduce delay along edges.
%\subsubsection{Remark 2.}
% -----------------------------------
%\textcolor{blue}{Mac: For the sake of symmetry, please comment on the computation of this integral here, just like in the first case.  Also, it might make sense to have a single theorem and proof that applies to both the node and edge computations.}

\subsubsection{Remark}

\vspace{2mm}
In this paper, we assume that the random variables $\CumulativeDelay{1}$, $\CumulativeDelay{2}$, $\DelayAtNode{v}{1}$, and $\DelayAtNode{v}{2}$ are mutually independent, which may not hold in applications where disturbances are likely to affect multiple robots at the same location. However, we can extend our approach to specific models in order to capture these effects, either by deriving new expressions for conflict probabilities and integrating them numerically, or simply by using Monte Carlo simulations.

In fact, we will show that if $\DelayAtNode{v}{1}$ and $ \DelayAtNode{v}{2}$ are not independent (but remain independent of $\CumulativeDelay{1}$ and $\CumulativeDelay{2}$), the probability of conflict at a node remains the same as long as the marginal distributions of $\DelayAtNode{v}{1}$ and $ \DelayAtNode{v}{2}$ are preserved.

%The above assumption is not unreasonable, as while the robots are likely to be similarly affected if they traverse a single location at a certain time, the delay that they have accumulated in the past can be considered independent from the delay that they will encounter in the future. 

Recalling the definition of events $A$ and $B$ in the case of a node conflict,

\begin{equation}
    \begin{aligned}
    P(\text{Conflict})
    &= P(A\cap B) = P(A) + P(B) - P(A\cup B) \\
    &= P(A) + P(B) - 1
    \end{aligned}
\end{equation}

\noindent
because we know that at least one of the events $A$ or $B$ will occur. Thus,

\begin{equation}
    \begin{aligned}
    P(\text{Conflict})
    &= \int_\mathbb{R} \big( P(A \mid y) + P(B \mid y) \big) P(y)dy - 1.
    \end{aligned}
\end{equation}

Using the previous expressions for the conditional distributions of $A \mid y$ and $B \mid y$ (derived from the marginal distributions of $\DelayAtNode{v}{1}$ and $\DelayAtNode{v}{2}$), we can write the sum

\begin{equation}
    \begin{aligned}
    P(A \mid y) + P(B \mid y)
    &= 1 + \int_{\vert \NominalArrivalTime{1} - \NominalArrivalTime{2} + y \vert}^{\infty} G(x \mid n_{\Node}, \lambda )dx.
    \end{aligned}
\end{equation}

We finally recover the previous expression for $P(\text{Conflict})$, 

\begin{equation}
\label{eq:NodeConflict2}
    \begin{aligned}
    P(\text{Conflict})
    &= \int_\mathbb{R} \int_{\vert \NominalArrivalTime{1} - \NominalArrivalTime{2} + y \vert}^{\infty} G(x \mid n_{\Node}, \lambda )P(y)dxdy.
    \end{aligned}
\end{equation}

Notice that we did not need to assume the conditional independence of $A$ and $B$ given $y$. This means that the expression for a node conflict from \cref{nodeprobability} holds even if we assume that two robots traversing a certain node at the same time will be affected by the same disturbances.

%\textcolor{blue}{Mac: This again is not enough detail.  If we want to claim this as a contribution, we need to expand on this claim and give the details. This remark should comment on the intuition and reasonableness of this assumption, not just state the assumption.  You want to give the reader some confidence that this isn't pulled out of thin air, that it is no unreasonable to expect this independence.}

% ------------------------------------------------------ %

% -------------------- ALGORITHM ----------------------- %
\section{Algorithm Description}\label{sec:algorithm}

%\subsection{Conflict-Based Search: Algorithm and Heuristics}

\newcommand{\ConstraintTreeNode}{\mathcal{P}}
\newcommand{\solution}{s}
\newcommand{\optimalsolution}{s^\star}

\subsection{STT-CBS}

STT-CBS uses the same high-level structure as CBS, reproduced in \cref{cbsalg}. Our main contributions reside in conflict detection and resolution.
The stochastic travel time model enables the algorithm to operate on simpler graphs where decomposition of time into time steps is not required. More specifically, nominal travel times at edges (or edge weights) can be arbitrary, in contrast with methods that constrain edge travel time to integers or unit time.

%\textcolor{blue}{Mac: This is the first time we have stated this claim, which is not good.  This should be clearly stated as a claim in the intro, or not mentioned at all.  It seems a bit dubious to me.  I would just not mention this as it may provoke fruitless debate with the reviewers.} - OP: section added at the end of the intro

\begin{algorithm}
 \caption{High level of CBS and STT-CBS}
 \begin{algorithmic}[1] % <-- \begin{algorithmic}[<starting_line_number>] 
 \renewcommand{\algorithmicrequire}{\textbf{Input:}}
 \renewcommand\algorithmicthen{}
 \renewcommand\algorithmicdo{}
%   \PROCEDURE{CBS}{}
  \STATE \textit{Initialize root node $\mathcal{R}$}:
  \STATE $\mathcal{R}$.constraints $\leftarrow$ $\emptyset$
  \STATE $\mathcal{R}$.solution $\leftarrow$ Find low level solution to $\mathcal{R}$ using A*
  \STATE $\mathcal{R}$.cost $\leftarrow$ Find solution cost
  \STATE Insert $\mathcal{R}$ into priority queue
  \WHILE{Priority Queue $\neq \emptyset$}
  \STATE $\mathcal{P}$ $\leftarrow$ best node in Priority Queue
  \STATE $\mathcal{P}$.conflict $\leftarrow$ GetFirstConflict($\mathcal{P}$) \label{eqn:get_conflict_line}
  \IF {$\mathcal{P}$ is conflict-free}
  \STATE \textbf{return} $\mathcal{P}$.solution
  \ENDIF
  \FOR{each agent $a_i$ involved in $\mathcal{P}$.conflict.agents}
  \STATE Create child node $\mathcal{C}$
  \STATE $\mathcal{C}$.constraints $\leftarrow$ $\mathcal{P}$.constraints + Constraint-From-Conflict($\mathcal{P}$.conflicts, $a_i$)
  \STATE $\mathcal{C}$.solution $\leftarrow$ Find low level solution to $\mathcal{C}$ using A*
  \STATE $\mathcal{C}$.cost $\leftarrow$ Find solution cost
  \STATE Add $\mathcal{C}$ to priority queue
  \ENDFOR
  \ENDWHILE
%   \ENDPROCEDURE
 \end{algorithmic} 
 \label{cbsalg}
 \end{algorithm}
In the original CBS algorithm, conflicts are detected by searching all pairs of robots at each time step. In STT-CBS, we search each node and edge containing occupants that could conflict. For example, for each occupied edge, we consider all combinations of robots that, under stochastic delays, could possibly traverse in opposite directions at the same time.
\Cref{fcalg} details the conflict detection process. 
%\textcolor{blue}{I am not clear on how the the Algo shows this?  You should tell the reader specifically how this algo differs from that in CBS.  The reader won't know this already.} 
Similarly to CBS, we expand our constraint checking tree with the first conflict found. 
We note that while evaluating the conflict probability at an edge, we search adjacent edges for occupancy information. When agents both occupy multiple adjacent edges in reverse directions, we compute the conflict probability for the longest sequence of these edges. This step is necessary to ensure that in the returned solution, the pairwise conflict probability between two agents (i.e. over their entire paths) is smaller than $\epsilon$.
For example, if agents $\Agent{1}$ and $\Agent{2}$ traverse the same sequence of edges $[e_1, e_2]$ in reverse directions, the conflict probabilities on $e_1$ and $e_2$ separately may each be smaller than $\epsilon$, although the actual conflict probability along $[e_1, e_2]$ may be larger than $\epsilon$. \Cref{fcalg} considers $[e_1, e_2]$ as a single edge and captures the actual possibility of a conflict occurring along the path.
The probability of a conflict along sequences of edges can be determined similarly to that of standard node and edge conflict probabilities, and should account for the delay at the nodes contained within the edges of the sequence. In our experiments, we use Monte Carlo Simulation to compute these probabilities.
%\textcolor{blue}{Mac: Sorry, I don't follow this.  Why does this logic only extend to pairs of edges?  Can't the same be said of triplets, etc, so the probability of collision along the whole path needs to be considered?  Can you please explain this a different way?  Is this also true for nodes, or only edges?  Why or why not?}
%In this case, checking for conflicts at each edge will not necessarily result in high conflict probabilities, whereas a probability computation for the entire path will indicate a conflict. 
%\textcolor{blue}{Mac: Sorry, also not getting the logical flow in this piece.  Are we saying without the look ahead to neighboring edges, we would incorrectly compute the collision probability?  Just not getting the picture (which means the reader will not either).}

\begin{algorithm}
 \caption{Get First Conflict}
 \begin{algorithmic}
 \renewcommand{\algorithmicrequire}{\textbf{Input:}}
 \renewcommand\algorithmicthen{}
 \renewcommand\algorithmicdo{}

 \STATE \textit{Fill the graph with occupancy information}:
  \FOR {each edge traversed by agents in $S$}
  \STATE Add occupancy information to the graph at nodes and edges: agent $R$, expected arrival time $t$, shape factor $n$
  \ENDFOR
% \STATE FirstConflict \gets $\emptyset$
  \FOR {each occupied vertex and edge in the graph}
  \FOR {each of their pairwise, possibly conflicting occupants}
  \IF{considered element is an edge}
  \STATE Search for adjacent edges traversed by both agents
  \STATE Combine these edges into a single edge
  \ENDIF
  \STATE Calculate $P_{conflict}$ using (\ref{nodeprobability}) or (\ref{edgeprobability})
  \IF {$P_{conflict} \geq \epsilon$}
  \STATE \textbf{return} Conflict(concerned agents, planned times of arrival)
  \ENDIF
  \ENDFOR
  \ENDFOR
 \STATE \textbf{return} None
 \end{algorithmic} 
 \label{fcalg}
 \end{algorithm}

\begin{algorithm}
 \caption{Constraint From Conflict}
 \begin{algorithmic}
 \renewcommand{\algorithmicrequire}{\textbf{Input:}}
 \renewcommand\algorithmicthen{}
 \renewcommand\algorithmicdo{}

 \STATE $\NominalArrivalTime{1} \gets$ time at which agent $a_i$ planned to arrive at the node or edge
 \STATE $n_1 \gets$ delay distribution parameter for agent $a_1$
  \STATE $\NominalArrivalTime{2} \gets$ time at which the other agent planned to arrive
 \STATE $n_2 \gets$ delay distribution parameter for agent $a_2$
 \STATE $P_{conflict} \gets 1.0$
 \STATE $k \gets 0$
 \WHILE{$P_{conflict} > \epsilon$}
 \STATE $k \gets k + 1$
 \STATE $P_{conflict} = P_{c\Node}(\NominalArrivalTime{1}+k\Delta t - \NominalArrivalTime{2},n_1,n_2,\lambda,n_\Node)$ or $P_{conflict} = P_{ce}(\NominalArrivalTime{1}+k\Delta t - \NominalArrivalTime{2},n_1,n_2,t_e,\lambda)$
 \ENDWHILE
 \STATE \textbf{return} Constraint($\Agent{1}$, element, $t_1+k\Delta t$)
 \end{algorithmic} 
 \label{cfcalg}
 \end{algorithm}

Given a conflict between $\Agent{1}$ and $\Agent{2}$, we create the constraint ``$\Agent{1}$ yields to $\Agent{2}$'' by finding the smallest necessary delay for $\Agent{1}$, denoted $t_{delay}$, that makes the conflict probability smaller than $\epsilon$. Then, the constraint prevents $\Agent{1}$ from entering the node or edge in question before time $\NominalArrivalTime{1} + t_{delay}$, where $\NominalArrivalTime{1}$ is the originally planned arrival time for $\Agent{1}$.
To compute this delay, we can use several possible algorithms, including linear search (\cref{cfcalg}) and binary search. In both cases, we will obtain an over-approximation of the minimal delay within some tolerance. 
The created constraints are then added to the set of constraints and incorporated into $A^\star$.

% To do this, we use binary search, as is explained in Algorithm \ref{dichalg}. We are able to use binary search, as the function $t_{delay} \mapsto P_{c\Node}(\NominalArrivalTime{1}+t_{delay}-\NominalArrivalTime{2},n_1,n_2,\lambda,n_\Node)$ is monotonous on the interval $[\NominalArrivalTime{1},\inf[$ in the case of nodes, and the same applies to $t_{delay} \mapsto P_{ce}(\NominalArrivalTime{1}+t_{delay}-\NominalArrivalTime{2},n_1,n_2,\lambda,t_e)$. The same reasoning applies for the second constraint, for which robot $\Agent{2}$ is delayed. The created constraints are then added to the set of constraints and incorporated into $A^\star$.}

% Original explanation for linear search
%Given a conflict between $\Agent{1}$ and $\Agent{2}$, one way to create the constraint ``$\Agent{1}$ yields to $\Agent{2}$'' is to calculate conflict probability with an increasing delay for $\Agent{1}$. Then the time the agent should yield is the smallest delay that yields a probability smaller than $\epsilon$. We use hyper-parameter $\Delta t$ in order to gradually increase delay. Until the returned time, the yielding agent is not authorized to use the element (node or edge). %This constraint is then added to the set of constraints and incorporated into $A^\star$. 
%Choosing a small value for $\Delta t$ costs the algorithm additional iterations. However, if $\Delta t$ is too large, we lose a key assumption that ensures the optimality of the returned solution. Empirically, we hand-tune this parameter according to our problem. 

\subsection{Properties}
We will prove that under our model, % as $\Delta t \xrightarrow{} 0$, 
the solution returned is optimal in terms of expected sum of travel times.
We will follow the same procedure as \citet{cbs}.

% Was erased for Wafr, but would have been useful
A key assumption is required: we suppose that in the unlikely event that two robots do conflict with each other, they do not accumulate additional delay. In other words, we ignore the effect of conflicts that do happen. By setting the threshold $\epsilon$ to a sufficiently small value, we can accept this assumption when the global conflict probability on our considered time horizon is small.
We do not know of a tractable method to reason about these conflicts and their propagation during planning. However, one effective way to compensate for the error made is to alter the parameters of the gamma distribution representing agent delay at each node such that it captures the additional delay that may be due to occurring conflicts. % We had removed this, but it seems that this particular point created confusion, so it is now rewritten and included again.
%\textcolor{blue}{Mac: Again, let's make the notation consistent here.  We haven't used $N$ for node earlier in the paper.}

\vspace{1mm}

\textbf{Definition 1} \textit{A solution $\solution$ is valid if and only if each pairwise conflict probability, at each node and longest edge traversed by opposing robots, is smaller than a threshold $\epsilon$.}
\vspace{1mm}

\textbf{Definition 2} \textit{For a given node $\ConstraintTreeNode$ in the constraint tree, we will call CV($\ConstraintTreeNode$) the set of valid solutions that do not violate constraints of $\ConstraintTreeNode$.}
\vspace{1mm}

\textbf{Definition 3}  \textit{A constraint tree node $\ConstraintTreeNode$ permits a solution $\solution$ if and only if $\solution$ $\in$ CV($\ConstraintTreeNode$).}
\vspace{1mm}

\textbf{Lemma 1} \textit{When creating a pair of constraints from a conflict with \cref{cfcalg}, we return the delay for each agent that brings the probability of conflict to $\epsilon$.}

%\textbf{Lemma 1} \textit{As $\Delta t \xrightarrow{} 0$, each resolved constraint ensures that the corresponding collision probability tends towards $\epsilon$.}

\begin{proof}

We prove properties of the continuous time function mapping delay time to collision probability that enable the sound application of linear search in \cref{cfcalg} to compute the optimal delay. We note that in practice, the delay computed will always differ from the optimal delay with some tolerance.

We will prove the statement in the case where we delay agent $\Agent{1}$. The proof extends to delaying the second agent $\Agent{2}$.

\item
\paragraph{Node constraints}
For fixed $\NominalArrivalTime{2}, n_1, n_2$, the function $f$ mapping $\NominalArrivalTime{1} \in \mathbb{R}^+$ to $f(\NominalArrivalTime{1}) = P_{c\Node}(\NominalArrivalTime{1}-\NominalArrivalTime{2}, n_1, n_2,\lambda, n_\Node)$ is a continuous function of $\NominalArrivalTime{1}$. We show that $f(\NominalArrivalTime{1}) \xrightarrow{} 0$ when $\NominalArrivalTime{1} \xrightarrow{} \infty$ (the same applies for $\NominalArrivalTime{2}$). 

Using the expression of node conflict probability in Section II, we will first show that
\begin{equation}
\label{infty_convergence_1}
\lim_{\NominalArrivalTime{1} \to \infty} \int_{0}^{\infty} P_{pos}(y)\int_{\mid \NominalArrivalTime{2} - \NominalArrivalTime{1} - y \mid}^\infty G(x\mid n_\Node, \lambda)dx dy = 0 .
\end{equation}

Let $(\alpha_n)_n$ be a sequence of real numbers, such that $\alpha_n~\to~\infty$ as $n~\to~\infty$. 
Let $(f_n)_n$ be the sequence of functions, such that  for all $n \in \mathbb{N}$ and $y \in \mathbb{R}^+$,
$$f_n(y) = P_{pos}(y)\int_{\mid \NominalArrivalTime{2} - \alpha_n - y \mid}^\infty G(x\mid n_\Node, \lambda)dx. $$

As $n \to \infty$, $f_n$ converges pointwise towards $0$. In addition, for all $n \in \mathbb{N}$ and $y \in \mathbb{R}^+$, $ |f_n(y)| \leq P_{pos}(y) $, where $P_{pos}$ is integrable on $\mathbb{R^+}$.
A standard application of the dominated convergence theorem yields
\[
\lim_{n \to \infty} \int_{0}^{\infty} f_n(y)dy = 0 .
\]

\noindent
Because the latter is true for any sequence $(\alpha_n)$, ~\eqref{infty_convergence_1} is proved.

The reasoning is identical for the second part of the expression.
\begin{equation}
\label{infty_convergence_2}
\lim_{\NominalArrivalTime{1} \to \infty} \int_{-\infty}^{0} P_{neg}(y')\int_{\mid \NominalArrivalTime{2} - \NominalArrivalTime{1} - y' \mid}^\infty G(x'\mid n_\Node, \lambda)dx' dy' = 0
\end{equation}

Therefore, from~\eqref{infty_convergence_1} and~\eqref{infty_convergence_2}, we obtain that for a node, $f(\NominalArrivalTime{1}) \to 0$ as $\NominalArrivalTime{1} \to~\infty$.

\item
\paragraph{Edge constraints}
For an edge and for fixed $\NominalArrivalTime{2}$, $n_1$, $n_2$, $t_e$, and $\lambda$, we denote $g$ the function mapping $\NominalArrivalTime{1}$ to $P_{ce}(\NominalArrivalTime{2}-\NominalArrivalTime{1},n_1,n_2,\lambda,t_e)$. Let us also define $(\alpha_n)_n$ a sequence of real numbers such that $\alpha_n \to \infty$ as $n \to \infty$.
Let $(g_n)_n$ be the sequence of functions such that $ \forall n \in \mathbb{N}, \forall x \in \mathbb{R}^+$:
$$
g_n(x) = G(x \mid n_1,\lambda) \int_{(h_n(x)-t_e)^+}^{(h_n(x)+t_e)^+} G(w \mid n_2, \lambda)  dw 
$$
with $$h_n(x) = x +\NominalArrivalTime{2}-\alpha_n$$
$(g_n)_n$ converges pointwise towards $0$.
Let us define $u$ such that $\forall x \in \mathbb{R^+}$, $u(x) = G(x \mid n_1, \lambda)$.
Then $u$ is integrable on $\mathbb{R}^+$ and $\forall x \in \mathbb{R}^+, \mid g_n(x)\mid \leq u(x)$. Thus, the theorem of dominated convergence also applies, and we obtain $g(\NominalArrivalTime{1}) \to 0$ as $\NominalArrivalTime{1} \to \infty$.

\vspace{5mm}

Finally, as we take steps of size $\Delta t$ until the probability becomes smaller than $\epsilon$, then for any $\delta \in \mathbb{R}^{+*}$, we are able to find $\Delta t \in \mathbb{R}^{+*}$ and $k \in \mathbb{N}$ such that $P_{ce}(\NominalArrivalTime{2} - \NominalArrivalTime{1} - k\Delta t, n_1, n_2, \lambda, t_e) \in [\epsilon - \delta, \epsilon]$. The same applies for a node conflict, where we compute $k$ and $\Delta t$  such that $P_{c\Node}(\NominalArrivalTime{2} - \NominalArrivalTime{1} - k\Delta t, n_1, n_2, \lambda, n_\Node) \in [\epsilon - \delta, \epsilon]$. Therefore, we have proved Lemma 1. %\qed

\end{proof}

% Include this, or not include this?
\textbf{Remark}
We did not show that Lemma 1 holds when computing the optimal delay using binary search in lieu of linear search. To complete such a proof, it would be necessary to show that $f$ and $g$ are uni-modal functions of $\NominalArrivalTime{1}$. If so, we would be sure to search for the optimal delay along a monotonously decreasing function,  which would guarantee that Lemma 1 holds. However, it is non-trivial to demonstrate whether $f$ and $g$ are (or are not) uni-modal, and we do not attempt to do so here.

\vspace{5mm}

%\vspace{-5pt}
\textbf{Lemma 2} \textit{Let $\ConstraintTreeNode$ be a constraint tree node with a non-empty set of constraints, an optimal (but not valid) solution $\optimalsolution \in CV(\ConstraintTreeNode)$ returned by $A^\star$, and children $\ConstraintTreeNode_1$ and $\ConstraintTreeNode_2$. Let $\solution$ be a valid solution. Then if $\solution \in CV(\ConstraintTreeNode)$, then at least one of the following is true: $\solution \in CV(\ConstraintTreeNode_1)$ or $\solution \in CV(\ConstraintTreeNode_2)$ }

\begin{proof} 
Let $\NominalArrivalTime{1}$ and $\NominalArrivalTime{2}$ be the planned times of conflicting agents $\Agent{1}$ and $\Agent{2}$ at the element, node or edge, causing a conflict in $\optimalsolution$ and the creation of $\ConstraintTreeNode_1$ and $\ConstraintTreeNode_2$. Since $\solution \in CV(\ConstraintTreeNode)$, we know that none of $\ConstraintTreeNode .constraints$ are violated. We know that $\Agent{1}$ cannot arrive at the element before $\NominalArrivalTime{1}$, and $\Agent{2}$ cannot arrive before $\NominalArrivalTime{2}$. Let $t_{D1}$ be the delay for agent $\Agent{1}$ that brings the conflict probability to $\epsilon$ while agent $\Agent{2}$ remains on its shortest path, and $t_{D2}$ the delay for agent $\Agent{2}$ that brings the conflict probability to $\epsilon$ while agent $\Agent{1}$ remains on its shortest path. 
 
 We know that we are able to find such times by using Lemma 1. More precisely, for a node, with $\delta = \NominalArrivalTime{1} - \NominalArrivalTime{2}$, we know that $P_{c\Node}(\delta,n_1,n_2,n_\Node,\lambda) \to 0$ as $\delta \to \infty$, and $P_{c\Node}(\delta,n_1,n_2,n_\Node,\lambda)~\to~0$ as $\delta \to -\infty$. The same applies for an edge.
 
 Then, we can show that there is no valid solution for which planned times for $\Agent{1}$ and $\Agent{2}$ will each belong to $[\NominalArrivalTime{1},t_{D1}[$ and $[\NominalArrivalTime{2},t_{D2}[$, respectively. In other words, a solution with such planned arrival and departure times has a conflict probability greater than $\epsilon$. We can easily state this with the methodology we used in order to find $t_{D1}$ and $t_{D2}$: we advance of steps  size $\Delta t$ until we find a conflict probability smaller than $\epsilon$. Thus, we know that when $\Delta t \rightarrow 0$, if $t_{D2}-(\NominalArrivalTime{1}-\NominalArrivalTime{2}) < \delta < t_{D1}-(\NominalArrivalTime{1}-\NominalArrivalTime{2})$, the corresponding conflict probability is strictly greater than $\epsilon$. Finally, because we cannot choose planned arrival times that are inferior to $\NominalArrivalTime{1}$ and $\NominalArrivalTime{2}$ for both agents, we can conclude that one of the new arrival times planned for agents $\Agent{1}$ and $\Agent{2}$ needs to be larger than $t_{D1}$ or $t_{D2}$, respectively. Therefore, for a valid solution, at least one of the additional constraints for planned arrival time is verified.
 
 %If we are to search for $t_{D1}$ and $t_{D2}$ using a more efficient method such as binary search, the Lemma still holds and the proof includes demonstrating the unimodality of $\delta \mapsto P_{c\Node}(\delta,n_1,n_2,n_\Node,\lambda)$. %\qed
 \end{proof}

\textbf{Lemma 3} \textit{The path returned by $A^\star$ for a given constraint tree node $\ConstraintTreeNode$ is a lower bound on the minimum cost of an element in $CV(\ConstraintTreeNode)$}

\begin{proof}
$A^\star$ returns the solution $S$ verifying all constraints of $\ConstraintTreeNode$, and minimizing the sum of expected travel times, which is the same as the expected sum of travel times. Therefore, the cost of $S$ is a lower bound on the cost of solutions that do not violate constraints of $\ConstraintTreeNode$. However, we know that $CV(\ConstraintTreeNode)$ is a subset of this set. Therefore, the cost of $S$ is a lower bound on the minimum cost of an element in $CV(\ConstraintTreeNode)$. %\qed
\end{proof}

\textbf{Lemma 4} \textit{Let p be a valid solution. At all time steps there exists a node $\ConstraintTreeNode$ in the priority queue that permits p.}

\begin{proof} 
We reason by induction on the expansion cycle \citep{cbs}.

Base case: At first, the priority queue (called \textit{OPEN} by \citet{cbs}) only contains the root node, which has no constraints. Consequently, the root node permits all valid solutions and also $p$.

Heredity: Let us assume this is true for the first $i$ expansion cycles, and call $\ConstraintTreeNode$ the concerned node in the priority queue that permits $p$. In cycle $i+1$, if node $\ConstraintTreeNode$ is not expanded, it remains in the priority queue, in which case the priority queue still permits $p$. On the other hand, let us assume that node $\ConstraintTreeNode$ is expanded and its children $\ConstraintTreeNode_1$ and $\ConstraintTreeNode_2$ are generated. Then, using Lemma 2, we can state that any valid solution for $\ConstraintTreeNode$ must be solution for $\ConstraintTreeNode_1$ or $\ConstraintTreeNode_2$. In both cases, there exists a node in the priority queue at the next expansion cycle that permits $p$.

Conclusion: For any valid solution $p$, at least one constraint tree node in the priority queue permits $p$.
By extension, there is always a node in the Priority Queue that contains the optimal solution. %\qed
\end{proof}

\textbf{Theorem 1} \textit{%As $\Delta t \xrightarrow{} 0$, 
If STT-CBS returns a solution, then it is the optimal solution with respect to the expected cost that ensures each pairwise conflict probability is smaller or equal to $\epsilon$.}

\begin{proof} 
Let us consider that the algorithm returns a valid solution $g$ from a goal node $G$. We know that at all times, all valid solutions are permitted by at least one node from the priority queue (Lemma 4). Let $p$ be a valid solution (with cost $c(p)$) and let $\ConstraintTreeNode(p)$ be the node that permits $p$ in the priority queue. Let $c(\ConstraintTreeNode)$ be the cost of node $\ConstraintTreeNode$. We have $c(\ConstraintTreeNode(p))$ $\leq$ $c(p)$ (Lemma 3). We know that, since $g$ is valid, $c(G)$ is a cost of a valid solution. Finally, similar to the work of \citet{cbs}, the search algorithm explores solution costs in a best-first manner. Due to this, we get that $c(g) \leq c(\ConstraintTreeNode(p)) \leq c(p)$, which proves Theorem 1. %\qed
\end{proof}

\vspace{1mm}

\textbf{Completeness in finite time approximation}
%The algorithm is sound, as shown in Theorem 1.
\Cref{cfcalg} ensures that each constraint delays a robot for a time that is larger than the positive hyper-parameter $\Delta t$. By doing so, we restrict the size of the search-space to a set with finite cardinality --- the set of arrival times for each agent at each node and edge where arrival times can only be delayed by multiples of $\Delta t$. This guarantees that STT-CBS will return the optimal solution within this set in a finite number of iterations.
 %However, we cannot guarantee that the algorithm will terminate, as the full state-space to explore through the search tree is a continuum, and the progress made at each node may be infinitely small. In other words, because the search space does not have fixed cardinality, there is no upper bound on the size of the search tree, and the algorithm may require an infinite number of iterations.

% Is there a preferred way to view the problem? 1- in theory, the algorithm is optimal, but not complete because the search space is too large. We cannot obtain that in practice as we cannot store exact real numbers. 2- In practice, we don't search the whole search space, because our delay is larger than some hyper-parameter (both with linear and binary search), so we may have missed out on an optimal solution. However, we return the best possible solution within this set. I'm not sure that we can say anything about how close the optimal solution within this set is to the optimal solution in general.
% ------------------------------------------------------ %

% --------------------EXPERIMENTS----------------------- %
\section{Simulations and Experiments}\label{sec:experiments}
\vspace{-5pt}
\begin{figure*}
\begin{minipage}{\textwidth}
\captionsetup[subfigure]{labelformat=empty}
\centering
\begin{subfigure}{.32\textwidth}
  \centering
  \includegraphics[width=\linewidth]{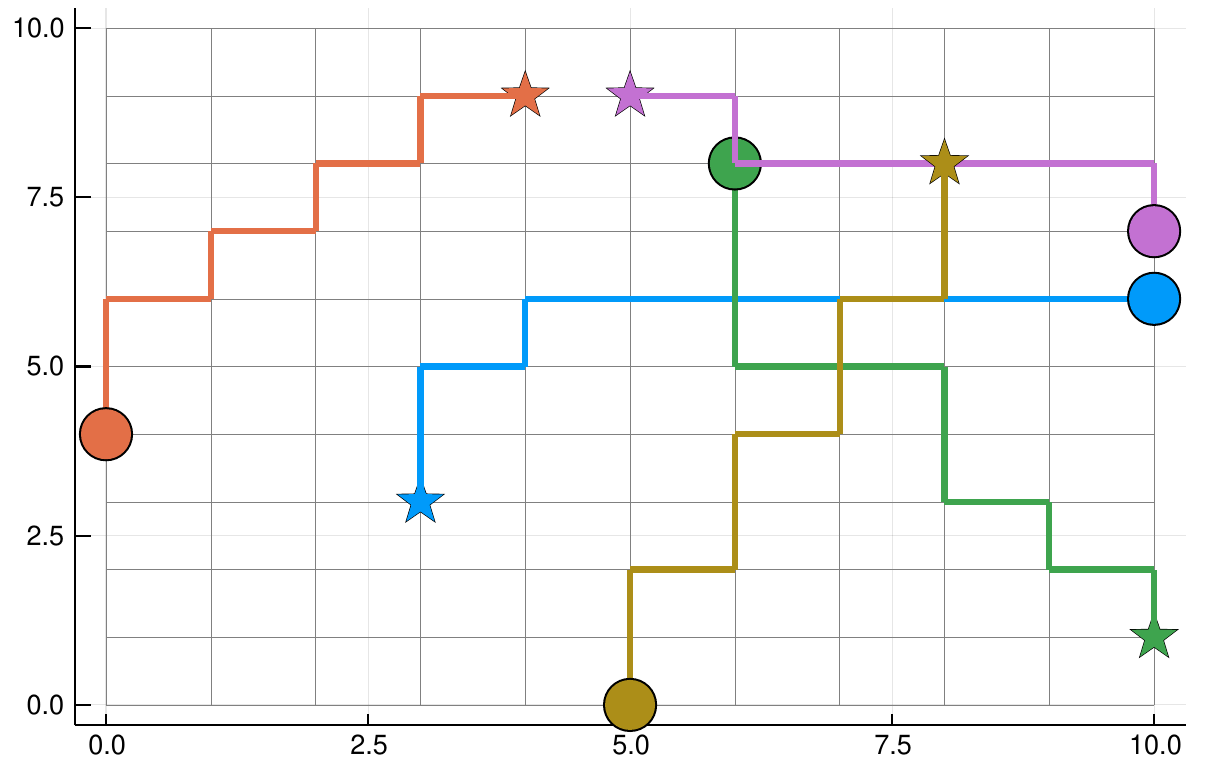}
  \caption{$\epsilon$ = 0.1}
  %\label{fig:sub1}
\end{subfigure}%
\begin{subfigure}{.32\textwidth}
  \centering
  \includegraphics[width=\linewidth]{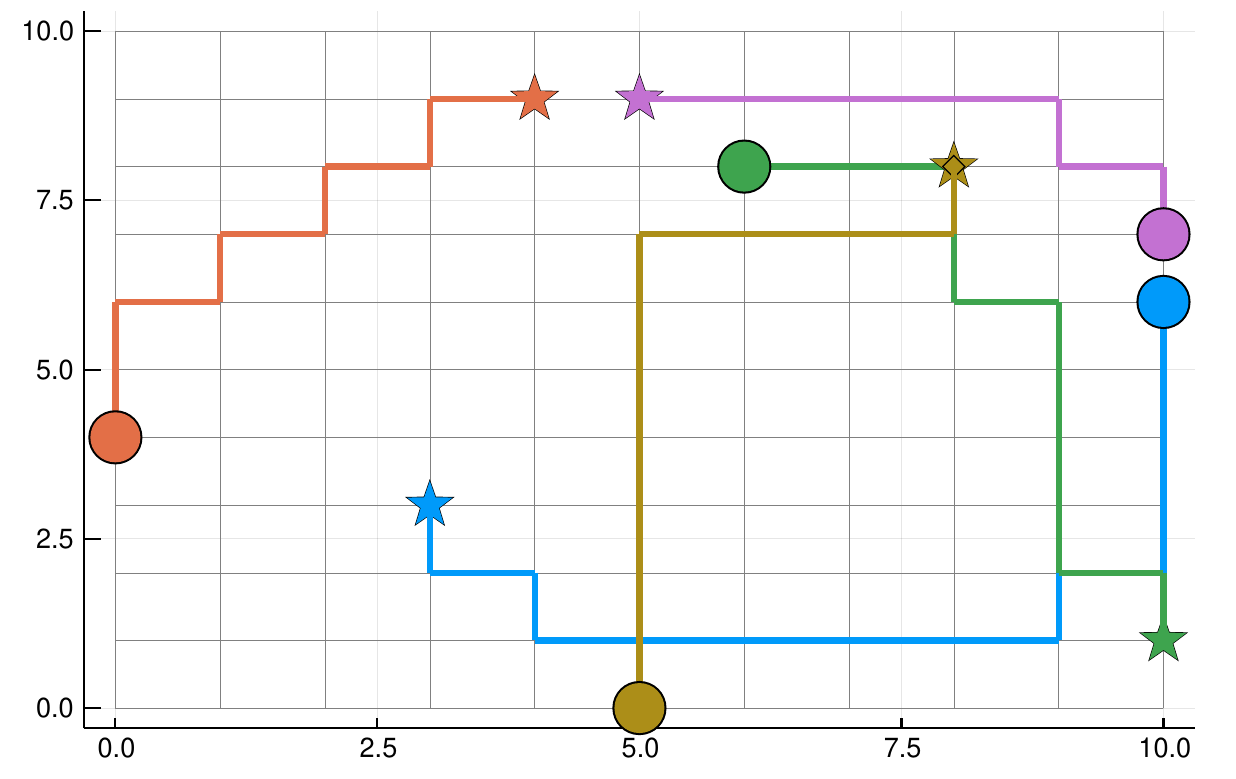}
  \caption{$\epsilon$ = 0.001}
  %\label{fig:sub2}
\end{subfigure}
\begin{subfigure}{.32\textwidth}
  \centering
  \includegraphics[width=\linewidth]{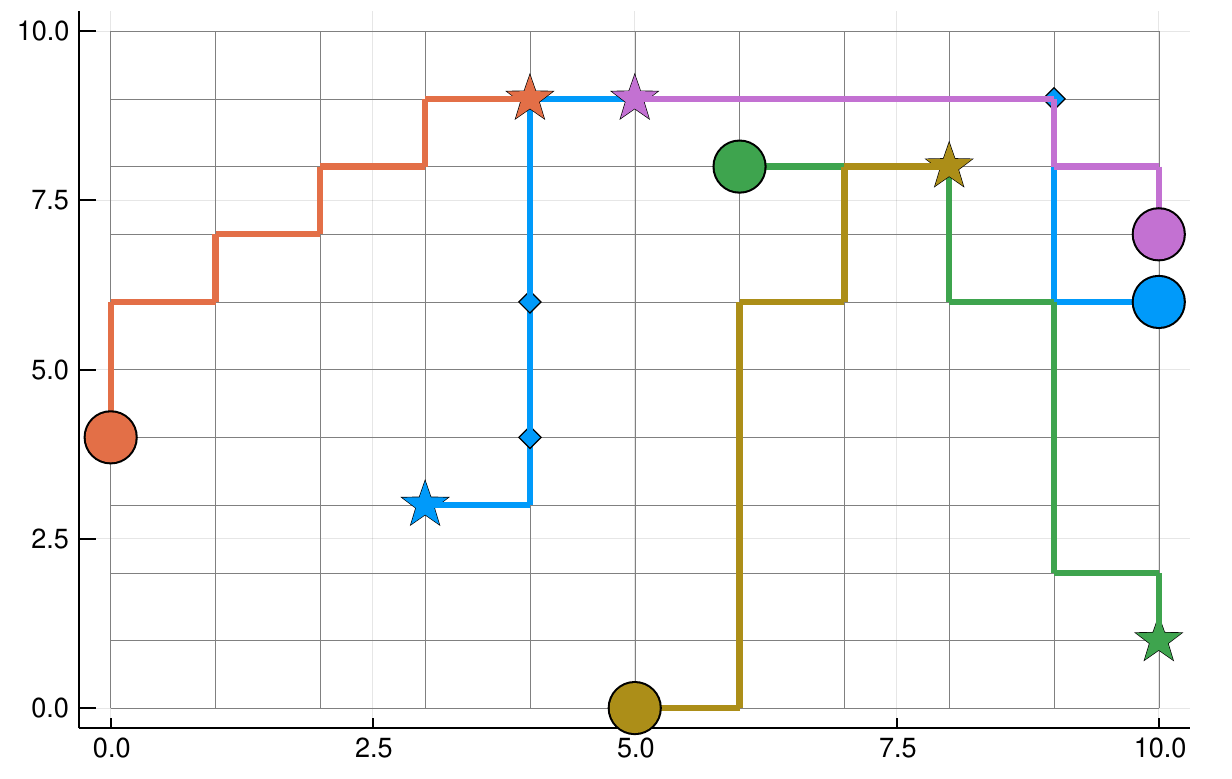}
  \caption{$\epsilon$ = 0.00001}
  %\label{fig:sub3}
\end{subfigure}
\caption{Solutions for STT-CBS on a $10 \times 10$ grid for 5 agents, with different values of $\epsilon$. Paths start at the stars, end at the circles, and are paused at the diamonds. With no uncertainty, each edge takes 1 time step to traverse.}
\label{fig:solutions}
\end{minipage}%
\end{figure*}

\begin{figure}
\centering
  \includegraphics[scale=0.4]{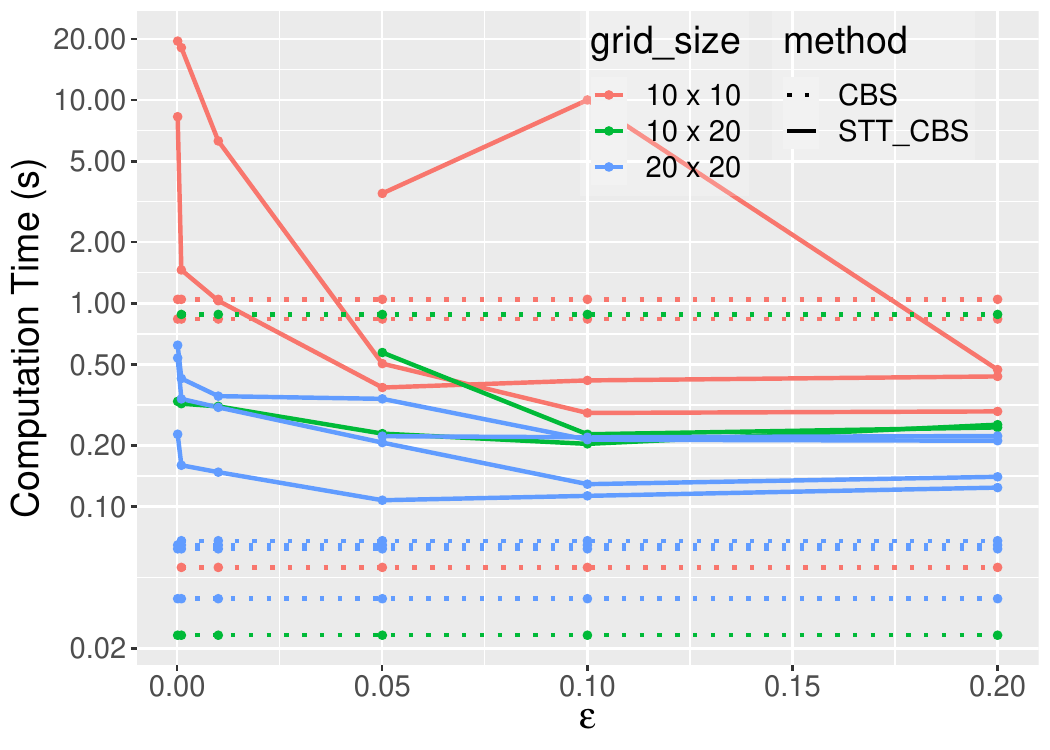}
  \caption{Computation time on a set of random examples for different values of $\epsilon$ on problems with 10 agents and three different grid sizes (time is in logarithmic scale). The dotted lines represent computation times for CBS applied to the same problems. The effect of reducing $\epsilon$, thus increasing problem complexity, is apparent. Whenever $\epsilon$ is sufficiently large, STT-CBS returns the optimal solution within one second. However, the extra control over collision probability comes with a moderately higher computation cost for our STT-CBS, compared with standard CBS.}
  \label{fig:compute_time}
\end{figure}

% \begin{figure*}
% \centering
% \begin{minipage}{.44\textwidth}
%   \centering
%   \includegraphics[scale=0.32]{graphics/conflict_probability.pdf}
%   \captionof{figure}{Global conflict probability evaluated for a number of agents varying between 8 and 12. We observe in practice that this value decreases as $\epsilon$ goes to 0.}
%   %\textcolor{red}{MS: Tell the view what they should get from the fig.}
%   \label{fig:globalcp}
% \end{minipage}%
% \begin{minipage}{.05\textwidth}
%   \hspace{1mm}
%   \vfill
% \end{minipage}
% \begin{minipage}{.44\textwidth}
%   \centering
%   \includegraphics[scale=0.32]{graphics/conflict_comparison.pdf}
%   \captionof{figure}{Global conflict probability comparison for randomly selected 10-agent problems. Each color represents a specific trial. Our algorithm gives the user a desired maximum pairwise collision probability $\epsilon$.  The conflict probabilities found with CBS are significantly larger than those found with our STT-CBS algorithm for small values of $\epsilon$. The global conflict probability is tied to local pairwise conflict probability (conditioned by $\epsilon$) through the number of interacting agents.}
%   %\textcolor{red}{MS: Tell the view what they should get from the fig.}
%   \label{fig:exp_com}
% \end{minipage}
% \end{figure*}

\begin{figure*}
  \centering
  \includegraphics[scale=0.4]{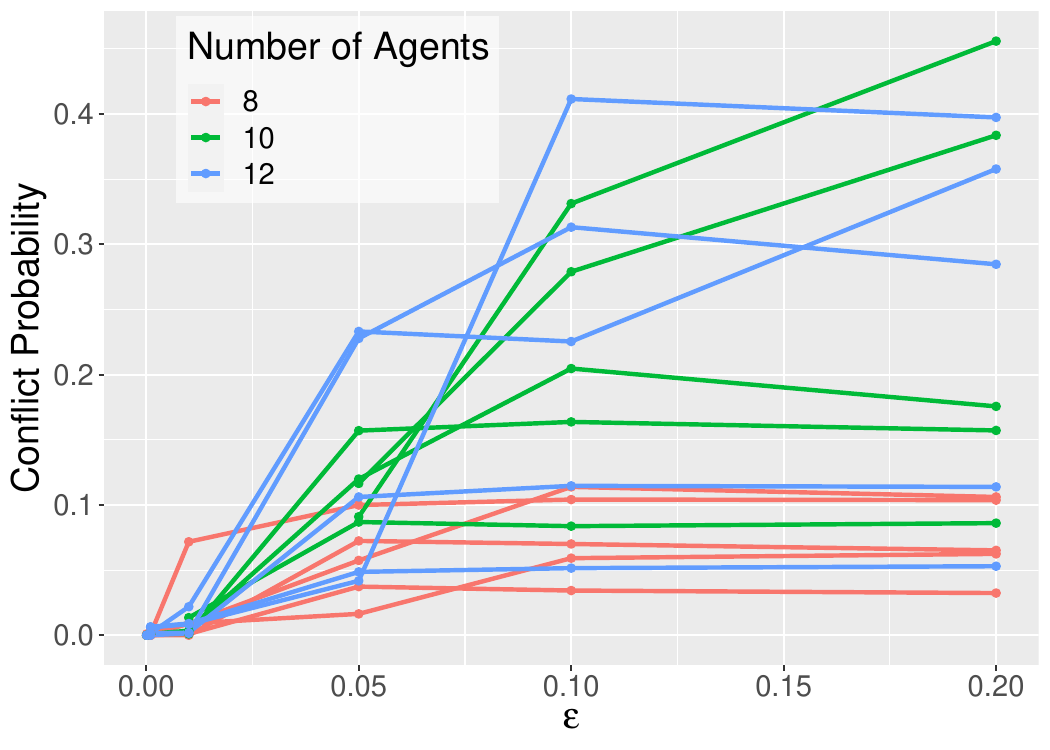}
  \captionof{figure}{Global conflict probability evaluated for a number of agents varying between 8 and 12. We observe in practice that this value decreases as $\epsilon$ goes to 0.}
  %\textcolor{red}{MS: Tell the view what they should get from the fig.}
  \label{fig:globalcp}
\end{figure*}

\begin{figure*}
  \centering
  \includegraphics[scale=0.32]{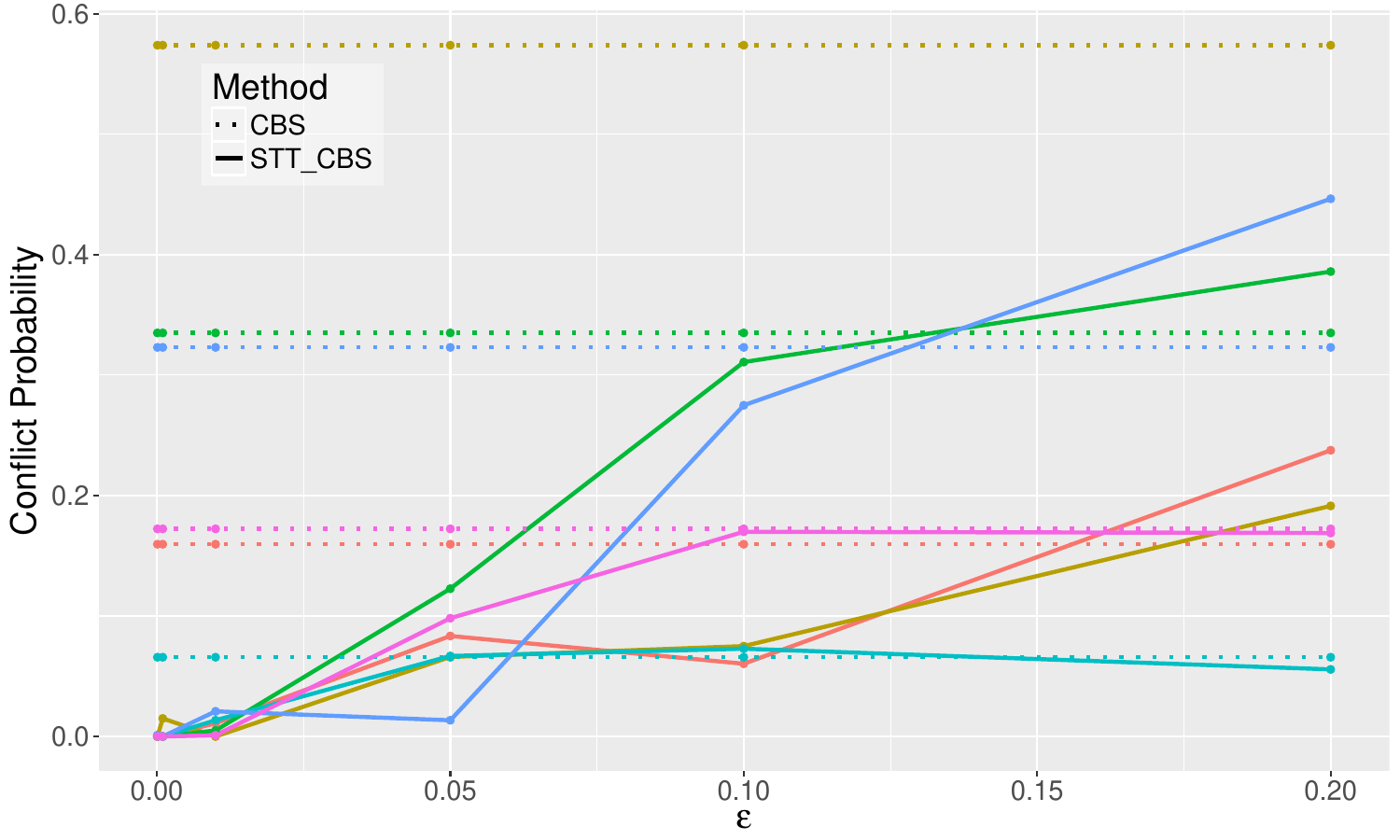}
  \captionof{figure}{Global conflict probability comparison for randomly selected 10-agent problems. Each color represents a specific trial. Our algorithm gives the user a desired maximum pairwise collision probability $\epsilon$.  The conflict probabilities found with CBS are significantly larger than those found with our STT-CBS algorithm for small values of $\epsilon$. The global conflict probability is tied to local pairwise conflict probability (conditioned by $\epsilon$) through the number of interacting agents.}
  %\textcolor{red}{MS: Tell the view what they should get from the fig.}
  \label{fig:exp_com}
\end{figure*}

%We evaluated STT-CBS against CBS in grid environments. 
\Cref{fig:solutions} illustrates the effect of $\epsilon$ on the returned solution, for an experiment on a $10 \times 10$ grid containing 5 agents. A high value yields a solution very likely to contain conflicts, and if the value is too small,  the solution tends to be over-conservative at the expense of a more costly solution. These results are also conditioned by the parameters of the delay distribution of every node.

For all the following experiments, we chose to model delay identically at every node by using parameters $\lambda=5$ and $n_\Node=1$.

\vspace{-5pt}
\paragraph{Computation time}
\Cref{fig:compute_time} shows computation time results for CBS and STT-CBS with different values of $\epsilon$. These experiments were performed on $10 \times 10$, $10 \times 20$ and $20 \times 20$ grids and with 10 agents. Whenever the number of iterations surpassed 1000, we interrupted the algorithm and did not represent the corresponding data point in these graphs.
It is important to note that while most of these randomly generated problems can be solved within a few seconds, due to the inherent difficulty of the problems solved with each method, neither CBS or STT-CBS provide any computation time guarantees.

\vspace{-5pt}
\paragraph{Conflict Probability}

\Cref{fig:globalcp} shows the global conflict probability on a set of experiments as a function of $\epsilon$. We computed the probabilities using Monte Carlo simulation. In these experiments, decreasing $\epsilon$ decreases global conflict probability. \Cref{fig:exp_com} compares the results with CBS. 
% The conflict probabilities found with CBS are significantly larger than those found with STT-CBS and reasonably large values of $\epsilon$. The global conflict probability is tied to local pairwise conflict probability (conditioned by $\epsilon$) through the number of interacting agents. % now in Figure caption
\paragraph{Hardware experiments}

\begin{figure}
\centering
  \includegraphics[scale=0.55]{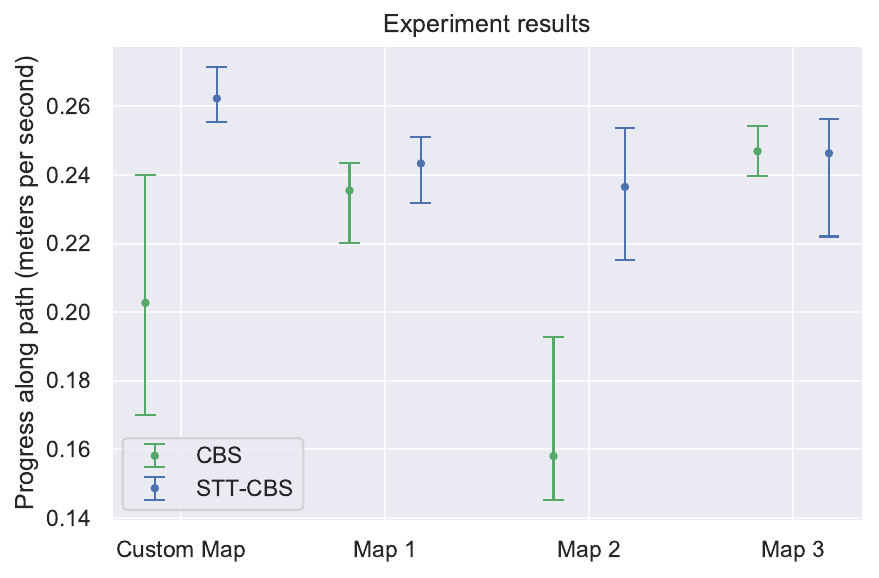}
   \caption{Hardware experiments. For each map and algorithm, 4 trials were conducted. The plot shows the average progress of the robots along their path, in meters per second. This result was obtained by dividing the sum of path lengths by the sum of travel times for all agents. The colored dots represent the mean, and the error bars represent the minimum and maximum sampled values. The collision avoidance algorithm was called while robots executed the CBS solution path in both the Custom Map and Map 2 due to robots navigating too closely to each other, resulting in slower progress over these trials. We find that that the additional safety provided by STT-CBS prevented such events to occur. The modeling choices we made in these experiments resulted in slightly longer path lengths, in return for gained predictability in travel time.}
%   \caption{Hardware experiments. For each map and algorithm, 4 trials were conducted. The colored dots represent the mean. The error bars represent the minimum and maximum sampled values. Travel times are in seconds. The collision avoidance algorithm was called while robots executed the CBS solution path in both the Custom Map and Map 2 due to robots navigating too closely to each other, resulting in a higher maximum travel time over these trials. We find that that the additional safety provided by STT-CBS prevented such events to occur. The modeling choices we made in these experiments resulted in slightly longer path lengths, in return for gained predictability in travel time.}
  % \textcolor{red}{MS: compared to what?  I only see STT-CBS on the plot. MS: What is the read supposed to get form this?  The caption should say: notice STT-CBS does a better job at XYZ.  Also, why only 4 trial?  Can we do 100 or 1000 trials?}
  % pieve of draft -- and the observed costs to vary significantly less, and consistently approach the nominal cost.
  \label{fig:exp_res}
\end{figure}

\begin{figure*}
\begin{minipage}{1.\textwidth}
\captionsetup[subfigure]{labelformat=empty}
\centering
\begin{subfigure}{1.\textwidth}
  \centering
  \includegraphics[width=\linewidth]{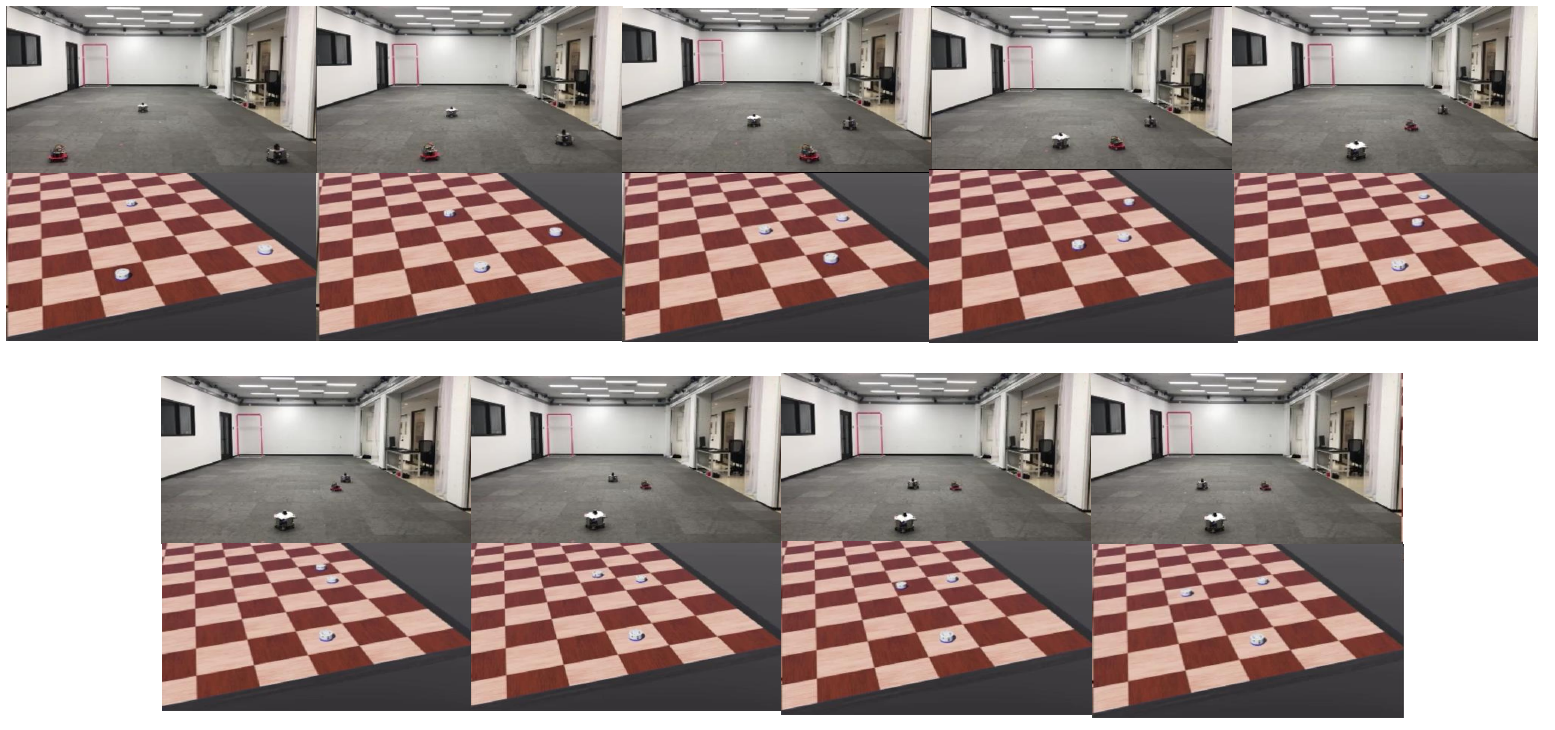}
  \caption{a) STT-CBS}
  %\label{fig:sub1}
\end{subfigure}%
\vspace{3mm}
\\
\begin{subfigure}{1.\textwidth}
  \centering
  \includegraphics[width=1.\linewidth]{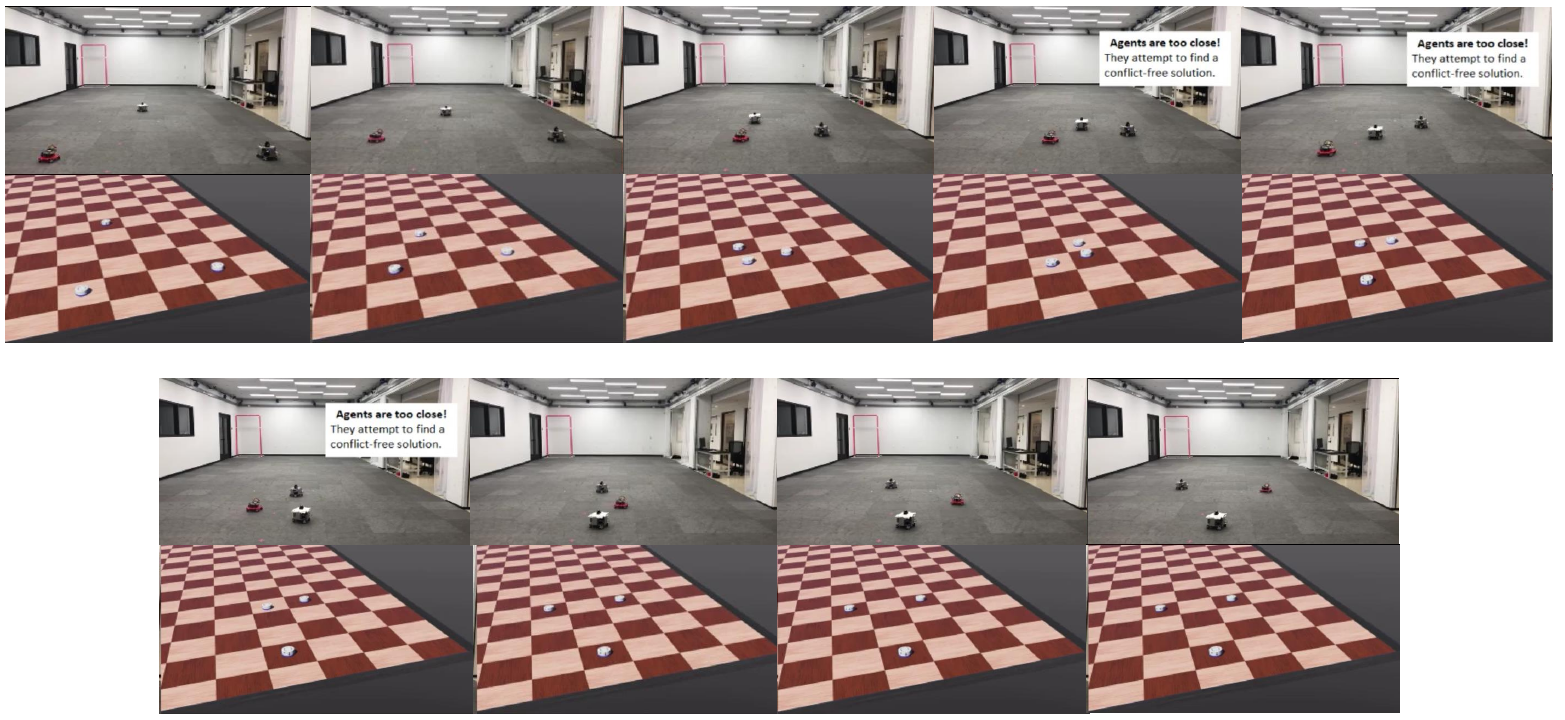}
  \caption{b) CBS}
  %\label{fig:sub2}
\end{subfigure}
\caption{We compare two different solutions computed by STT-CBS and CBS for the same problem in a $5 \times 5$ custom environment. At each time step, we compare agents' actual (upper) and simulated (lower) positions. We notice that the solution returned by CBS requires the agents to navigate close to one another as they follow each other. During the execution, agents are not able to perform this maneuver without risking collision and compute alternate routes to avoid each other. In contrast, STT-CBS returns a slightly more expensive solution that the agents are able to execute safely.}
\label{fig:multipane}
\end{minipage}%
\end{figure*}

In order to test our algorithm, we use the OuijaBot \citep{ouijabots}---a custom-designed omnidirectional platform---to execute solutions. To avoid collisions, we plan trajectories using Reciprocal Collision Avoidance for Real-Time Multi-Agent Simulation \citep{rvo2}.  In the experiments, three robots are instructed to follow way-points. Travel time is uncertain as their dynamics vary with time, battery level, and floor condition, and they may need to manoeuvre around each other. The distribution for agent delay is unknown and realistic.

We compare the performance of CBS and STT-CBS in one custom environment, and three chosen among 10 randomly generated environments where CBS and STT-CBS yield different solutions. We chose $n_\Node=1$ and $\lambda=5$ at all nodes to model delay, and $\epsilon=0.1$ as conflict probability threshold. The results are summarized in \Cref{fig:exp_res}. %, where the nominal cost was computed using the nominal robot velocity. 
%We find that that the additional safety provided by STT-CBS helps the observed costs to vary significantly less, and consistently approach the nominal cost (i.e. the cost with no stochastic delay in travel times).
\Cref{fig:multipane} illustrates a difficulty that becomes apparent when the computed path is not sufficiently robust to travel time uncertainty. We notice that the solution generated by CBS for the custom environment requires the agents to navigate close to one another as they follow each other. During execution, agents fail to perform this maneuver without risking collision and compute alternate routes to avoid each other. In contrast, STT-CBS returns a more expensive solution that the agents are able to execute safely.

% ------------------------------------------------------ %

% ------------------ CONCLUSION ------------------- %
\section{Conclusion}

STT-CBS offers quantifiable robustness to stochastic travel time delays in realistic multi-agent path finding scenarios compared with the standard CBS method, while minimizing the expected solution cost. 
An interesting direction for future work is the integration of stochastic travel time models into other path planning algorithms, such as some of the many variants and extensions of Conflict-Based Search \citep{cbsta,cbm,disjointcbs,kyle}.
In addition, the multi-agent path planning literature will benefit from a comparison of different `robust' MAPF solvers in both realistic simulations and hardware experiments. 

%While the original CBS algorithm outperforms STT-CBS in terms of computation time, we show that the the solutions generated by STT-CBS are more robust to uncertainty, both in simulation and with experiments. %This result assumes that our original model is accurate.

%The optimality property is only acceptable for small values of $\epsilon$. In future work, we will explore conflicts involving more than two robots in order to consider global conflict probability, which will benefit applications with larger values of $\epsilon$.
% ------------------------------------------------- %

\vskip 0.2in
\printbibliography

\end{document}